\documentclass[10pt,twocolumn,letterpaper]{article}

\usepackage{iccv}
\usepackage{times}
\usepackage{epsfig}
\usepackage{graphicx}
\usepackage{amsmath}
\usepackage{amssymb}

\usepackage{algorithm,algorithmic}
\usepackage{courier}
\usepackage{subfig}
\usepackage{enumitem}

\newtheorem{thm}{Theorem}

\usepackage[pagebackref=true,breaklinks=true,letterpaper=true,colorlinks,bookmarks=false]{hyperref}

\iccvfinalcopy 

\def\iccvPaperID{1827} 

\ificcvfinal\fi
\begin{document}

\title{A Tour of Convolutional Networks Guided by Linear Interpreters}

\author{Pablo Navarrete Michelini,\quad Hanwen Liu,\quad Yunhua Lu,\quad Xingqun Jiang\\
BOE Technology Co., Ltd.\\
{\tt\small \{pnavarre, liuhanwen, luyunhua, jiangxingqun\}@boe.com.cn}
}

\maketitle

\begin{abstract}
Convolutional networks are large linear systems divided into layers and connected by non--linear units. These units are the ``articulations'' that allow the network to adapt to the input. To understand how a network manages to solve a problem we must look at the articulated decisions in entirety. If we could capture the actions of non--linear units for a particular input, we would be able to replay the whole system back and forth as if it was always linear. It would also reveal the actions of non--linearities because the resulting linear system, a \textbf{Linear Interpreter}, depends on the input image. We introduce a hooking layer, called a \textbf{\mbox{LinearScope}}, which allows us to run the network and the linear interpreter in parallel. Its implementation is simple, flexible and efficient. From here we can make many curious inquiries: how do these linear systems look like? When the rows and columns of the transformation matrix are images, how do they look like? What type of basis do these linear transformations rely on? The answers depend on the problems presented, through which we take a tour to some popular architectures used for classification, super--resolution (SR) and image--to--image translation (I2I). For classification we observe that popular networks use a pixel--wise vote per class strategy and heavily rely on bias parameters. For SR and I2I we find that CNNs use wavelet--type basis similar to the human visual system. For I2I we reveal copy--move and template--creation strategies to generate outputs.
\end{abstract}

\section{Introduction}
\begin{figure}[t]
    \centering
    \includegraphics[width=\linewidth]{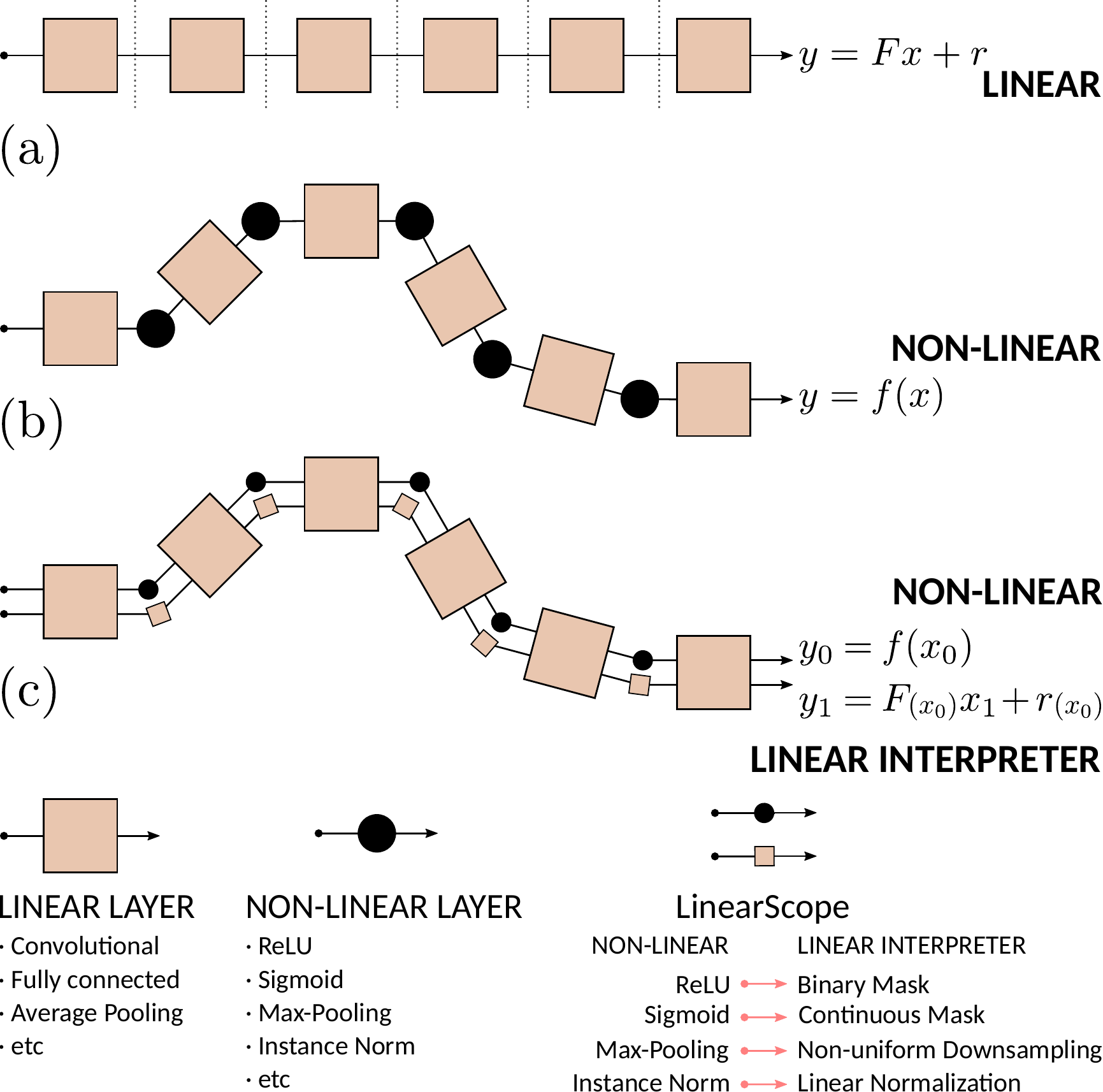}
    \caption{(a) Attaching linear layers of a network gives a linear system. (b) Non--linear units work as ``articulations'' that make the network adaptive to the input. (c) We can run the network in two batches, and use a LinearScope in each non--linear unit to run the network on the first batch, and a linear interpretation of the non--linear action in the second batch. The output in the first batch is unaffected by LinearScopes. The second batch gives a linear interpreter of the whole network that depends non--linearly on the first batch and linearly on the second batch.}
\label{fig:introduction}
\end{figure}
In this paper we are going to explore the interpretability of convolutional networks by using linear systems. The main task is to improve our understanding of how deep neural networks solve problems. This has become more intriguing because of the predominance of deep--learning systems in machine learning benchmarks, which has motivated extensive research in this area. The question is how to \emph{interpret} the models, and the interpretation can reflect many different ideas\cite{lipton2016mythos}. But before we get into the meaning of interpretability, let us first remember that the design of neural networks was simple from its very beginning: linear systems and (non--linear) \emph{activations}\cite{rosenblatt1958perceptron}. Here, \emph{activations} are biologically inspired to refer to inhibition of features (the output of the linear system), and the usual circuitry analogy is a switch. The problem arise when we combine many of these simple units, run many features in parallel, and subsequently repeating the same process. More precisely, it is not clear how the partial results lead us to the final decision.

Linear systems are generally considered \emph{interpretable} given a long history of research\cite{strang1993introduction}. With a linear system we know what to expect and where to look at to find answers. Here, we are interested in some of their most important properties. We will write an affine transformation $y:\mathbb{R}^n\rightarrow\mathbb{R}^N$ as
\begin{equation}
    y(x) = Fx + r
\end{equation}
where $r\in\mathbb{R}^N$ is a \textbf{residual} that lives in the same space as the output, and it is thus visible and interpretable as a fixed shift. The next useful information comes directly from the \textbf{rows} and \textbf{columns} of the matrix $F$. A row shows us the input pixels that are used to get an output pixel. We call these the \emph{receptive filter} coefficients as their extension in space show the \emph{receptive field} of the model. On the other hand, a column shows us the output pixels affected by an input pixel. We call them the \emph{projective filter} coefficients and their extension in space the \emph{projective field} of the model. Other important information comes from the \textbf{transposed system}, represented by $F^T$, which interchanges the meaning of rows and columns and \emph{back}--projects vectors from the output domain back to the input domain.

To interpret a linear transformation $Fx$ as a whole, which is to get a feeling of what parts of an input signal passes and how much it passes, we need its \textbf{singular value decomposition} (SVD), $F=U\Sigma V$. This gives us a full description of the vector spaces connecting input and output domains. The set of left ($U$) and right ($V$) eigenvectors basically shows us what the outputs and inputs are made of according to the transformation. For linear space invariant systems (LSI)\cite{JGProakis_2007a}, these are harmonic functions like complex exponentials $U_{jk}=e^{-i\Omega jk}$ or some type of DCT\cite{strang1999discrete}. These systems play a fundamental role in signal processing\cite{JGProakis_2007a,SMallat_1998a}. In simple terms, in LSI systems waves move in and out without changing their shapes, and can be interpreted as the natural choice of the system to decompose inputs and outputs. When a matrix is not symmetric or square, left and right eigenvectors are different. For the sake of simplicity, we will call them \textbf{eigen--inputs} and \textbf{eigen--outputs}, so that they remind us of the space where they live. What matters here is a pair of eigen--input, $v\in\mathbb{R}^n$, and eigen--output, $u\in\mathbb{R}^N$. An eigen--input transformed with $F$ gives us an eigen--output rescaled by its singular value $\sigma$, $u=\sigma Fv$, and the eigen--output back--projected with $F^T$ returns the rescaled eigen--input, $v=\sigma F^Tu$. So in general terms, a pair of eigen--input/output moves in and out, projected and back--projected, without changing their shapes, just rescaled by their singular values. The singular value shows the filtering effect, which represents what passes and how much passes. A small singular value indicates a pair of eigen--input/output that vanishes quickly after a transformation and back--projection.

Now, why should we use linear systems to interpret convolutional networks? We cannot study a structure made of material A by using our knowledge on material B, just because we know B better. Linearizations of convolutional networks can indeed be very useful, and have been studied in \cite{montavon2017explaining} to obtain heatmappings that show the relevance of inputs in the outputs. Its connections with our results will be discussed later. Here, we want to emphasize two simple arguments as to why should we use linear systems:
\begin{enumerate}[leftmargin=*]
    \item \textbf{Convolutional networks are largely made of linear systems}. In fact, all the parameters of a network are contained in linear modules (e.g. convolutional layers) with few exceptions (e.g. Parametric ReLU); \label{argument1}
    \item \textbf{The design of non--linear units have an initial linear motivation}, and the non--linearity is added in order to select their linear parameters adaptive to the input. Activations like ReLU or Sigmoid are switches that can be represented by pixel--wise masks multiplying inputs. If we fix the mask, it becomes linear. A max--pooling layer selects one among a group of pixels and allows a similar interpretation by using selection masks. An instance--normalization layer subtracts a mean and divides by a standard deviation. If we fix the mean and standard deviation, it becomes linear. Now, we do have simple linear interpretations of non--linear units.
\end{enumerate}
So, if we use the linear interpretation of non--linear layers (meaning to freeze the decisions of non--linear units), the whole system becomes linear. This procedure has been used in \cite{PNavarrete_2019a} to visualize how CNNs upscale small images. The authors proposed to replace activation units by masks and thus obtained linear systems of the form $y=Fx+r$. By inspecting the columns of $F$, they observed upscaling coefficients highly--adaptive to the input.

This work focuses on experimental explorations. Similar to a laboratory that needs a microscope to study microorganisms, we need an instrument to perform studies with linear interpreters. Thus, a key contribution is the design of a hooking layer (LinearScope), that can be inserted in CNNs to extract information. With this tool in hand we are able to extend an existing approach of interpretability\cite{PNavarrete_2019a} to significantly broader applications, through which we have made the following important discoveries:
\begin{itemize}[leftmargin=*]
    \item We report a \textbf{``pixel--wise vote'' interpretation of image classifiers} in which each pixel votes independently for an image label and the most voted label gives the output. Other works have found that classification CNNs are biased towards textures\cite{geirhos2018imagenet}, or that they still perform well after shuffling patches\cite{kang2017patchshuffle}, while our results point to the concrete strategy of the network (pixel votes).

    \item We report a \textbf{critical role of the bias parameters in CNNs for image classification}, as opposed to other applications (e.g. SR and I2I). Moreover, they become more relevant in architectures with better benchmarks and, in the case of sequential networks we find the contributions to concentrate on specific layers that move deeper when trained with batch normalization.

    \item We explain the \textbf{strategies of CycleGAN to solve I2I}. We uncover a copy--move strategy for photo--to--painting task (moving textures from different places in an image) and a template--creation strategy for the facades--to--label task. It should be noted that prior to this paper, it was largely unknown how to identify the source of newly generated objects and textures.

    \item We derive an algorithm using LinearScopes to \textbf{obtain the SVD of a linear interpreter}. This shows us the basis behind a CNN. Here, we found strong connections to the Human Visual System (HSV). It is known that the receptive fields of simple cells in mammalian primary visual cortex can be characterized as being spatially localized, oriented and bandpass, comparable to wavelet basis. In \cite{olshausen1996emergence} it is shown that a coding strategy that maximizes sparseness is sufficient to account for these properties, and have been of great impact in the field of sparse coding. Our SVD results reveal that the basis used by SR and I2I networks also contain all three properties above. In terms of output knowledge, it gives us an overview of the strategy to map input to output pixels.
\end{itemize}

These results may bring about the following \textbf{future impact}: 1) the explicit demonstration that CNNs use wavelet--type basis similar to the human visual system, 2) the creation of tools to visualize and fix problems in CNN architectures, and 3) the possibility to use the filter/residual in a loss function and design CNNs with an interpretable target.

\section{Related Work}
The interpretability of convolutional networks is closely related to visualization techniques. Visualization is more generally concerned on visual evidence of information learned by a network\cite{olah2017_featurevis}. Interpretability tries to explain the inner processing of a network, and each interpretation comes with a visualization technique that we can use to interpret the learning process. Reviews of the extensive literature in visualization can be found in \cite{zhang2018visual,qin2018convolutional,olah2018_buildingblocks,olah2017_featurevis}.

The meaning, or many meanings, of interpretability is a subject of study. In \cite{lipton2016mythos}, for example, authors identify a discordant meaning of interpretability in existing research and discuss the feasibility and desirability of different notions. They also emphasize an important misconception, that linear models are not strictly more interpretable than deep neural networks. In \cite{dhurandhar2017tip}, authors define interpretability relative to a target model and not as an absolute concept. In \cite{Adebayo47450}, authors show how assessments relying only on the visual appealing of saliency methods can be misleading and they propose a methodology to evaluate the explanations that a given method can provide. Finally, in \cite{ghorbani2017interpretation} authors show how the interpretation of neural networks is a fragile process, showing how they can introduce small perturbations in images leading to very different interpretations.

Extensive work has been done to explain the decisions of image classifiers and segmentation\cite{dhurandhar2018explanations,ribeiro2016should,DBLP:conf/icml/ShrikumarGK17,cnnfixations-mopuri-2017,balu2017forward,carletti2018understanding,du2018towards,dhurandhar2018explanations,ribeiro2016should,DBLP:conf/icml/ShrikumarGK17,cnnfixations-mopuri-2017,balu2017forward,carletti2018understanding,du2018towards,rottmann2018prediction}. Other research directions on image classification try to find answers inside a network architecture. In \cite{cadena2018diverse}, for example, authors study invariances in the responses of hidden units and find that these are major computational component learned by networks. In \cite{fong2018net2vec}, authors study the collaboration of filters to solve a problem and find that multiple filters are often required to code a concept, and single filters are not concept specific. In \cite{li2018decision}, authors show that the last layer of a network works as a linear classifier, similar to the motivation of the perceptron\cite{rosenblatt1958perceptron}.

An important research direction is to study the role of semantics. The Network--Dissection framework has been proposed in \cite{netdissect2017} to quantify the interpretability of latent representations by evaluating the alignment between individual hidden units and a set of semantic concepts. In \cite{zhou2018interpretable}, a new framework is proposed to decompose activations of the input image into semantically interpretable components. And the GAN--Dissection framework has been proposed for visualizing the structure learned by generative networks\cite{bau2019gandissect}.

Our interpretation of CNN--classifiers are more closely related to: \textbf{Layer--wise Relevance Propagation (LRP)}\cite{bach2015pixel,binder2016layer} and \textbf{Deep Taylor Decomposition} (DTD)\cite{montavon2017explaining}. LRP is the first framework to introduce a general solution to the problem of understanding classification decisions by pixel--wise decomposition of network classifiers, and DTD is the first study to consider Taylor decompositions in a network. The relation to our results will be discussed in Section \ref{sec:discussion}.

Finally, our analysis is an extension of \textbf{Deep Filter Visualization (DFV)}, introduced in \cite{PNavarrete_2019a} to visualize how convolutional networks upscale low--resolution images. DFV proposes to replace activation units by masks and thus obtains a linear system of the form $y=Fx+r$. DFV has been used to inspect the columns of $F$ and observe upscaling coefficients highly--adaptive to the input. In DFV one needs to record the activations for every non--linear unit in order to run the linear interpreter. This comes with a high storage cost for common architectures as shown in Table \ref{tab:storage}.  If we do not have enough memory in a device (e.g. GPU), we need to switch to slower storage such as CPU DRAM, SSD or HDD with an overwhelming cost in speed, as shown in Table \ref{tab:speed}. We propose a solution to this problem that does not require to store activations, and instead requires an additional batch in the input. This novel approach gives us a much simpler and efficient implementation of the linear interpreter. We are not only able to run faster and use larger images, but we can also perform more complex analysis on the linear interpreter, including: transposed linear interpreters and singular value decompositions. State--of--the--arts CNNs are often pushed to the limit of current technologies which makes our solution critical for experimental explorations with a $2\times$ to $10^4\times$ speedup over DFV\cite{PNavarrete_2019a} according to Tables \ref{tab:storage} and \ref{tab:speed}.
\begin{figure}[t!]
    \centering
    \includegraphics[width=\linewidth]{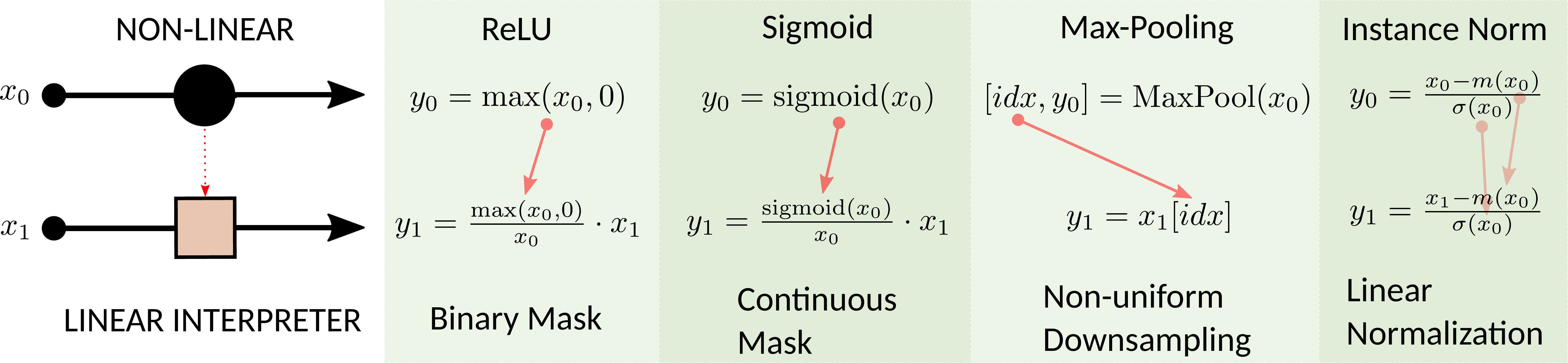}
    \caption{A LinearScope keeps a non--linear unit unchanged on batch $x_0$ and adds a second batch $x_1$ to run a linear interpreter. Red lines show how the interpreter looks at the first batch to decide: what mask to use (ReLU and Sigmoid), what inputs to select (MaxPooling), or what normalization mean and variance to use (Instance Normalization).}
\label{fig:linearscopes}
\end{figure}
\begin{table}
    \begin{centering}
    \begin{tabular}{l|ccc}
    \hline
    Network & VGG--$19$\cite{Simonyan14c} & CycleGAN\cite{CycleGAN2017} & EDSR\cite{Lim_2017_CVPR_Workshops} \\ \hline
    Space   &    $58$ GB &     $90$ GB & $4,147$ GB \\ \hline
    \end{tabular}
    \end{centering}
    \caption{Storage space needed to store all ReLU activations.} \label{tab:storage}
\end{table}

\begin{table}
    {\hfill
    \begin{tabular}{l|cccc}
    \hline
    Storage  & GPU & CPU &            SSD &            HDD \\ \hline
    Speed & $100\%$ & $50\%$ & $0.5\%$ & $0.005\%$ \\ \hline
    \end{tabular}\hfill
    }
    \caption{Relative speed of typical storage media, taking as reference GPU (DDR5 or HBM2).} \label{tab:speed}
\end{table}

\section{The Linear Interpreter}
\textbf{LinearScopes}: We define a LinearScope as a hooking layer that modifies a non--linear unit by adding an additional batch. If a non--linear unit calculates $y_0=h(x_0)$ on a batch $x_0$, then we change it to calculate:
\begin{equation}
    [y_0, y_1] = [h(x_0), A(x_0)\; x_1 + c(x_0)] \;.
\end{equation}
Here, $[\cdot,\cdot]$ denotes concatenation in the batch dimension, and $A(x_0), c(x_0)$ are chosen depending on our interpretation of $h(x_0)$. A hard requirement is
\begin{equation}
x_0 = x_1 \quad\Rightarrow \quad y_1=y_0 \;.
\end{equation}
One choice of linear interpreter is the best linear approximation of $h$ given by the Taylor expansion around the input:
\begin{align}
    h(x_1) & = h(x_0) + (Dh)(x_0) \cdot (x_1-x_0) + \cdots \\
           & = \underbrace{(Dh)(x_0)}_{A(x_0)} \cdot x_1 + \underbrace{h(x_0) - (Dh)(x_0) \cdot x_0}_{c(x_0)} + \cdots \nonumber
\end{align}
so that $y_1 = A(x_0)\; x_1 + c(x_0)$ is the Taylor interpreter.

Here, we follow and extend the approach of DFV\cite{PNavarrete_2019a}, which is not to seek an approximation. We prefer to use the word \emph{freezing} instead of \emph{linearization}. We think of the DFV approach as follows: the network has taken some decisions throughout its layers for an input image (See Figure \ref{fig:introduction}). Figure \ref{fig:linearscopes} shows the unique choices to fix these decisions. The overall \emph{frozen} system happens to be linear because of the particular structure of CNNs, as opposed to a Taylor expansion that forces linearity in the interpreter.

\textbf{Linear Interpreter}: Figure \ref{fig:introduction} explains our general idea. We want to use the LinearScope hooking layers inside a model to replace all its non--linear units. If a network outputs $y_0=f(x_0)$, with $x_0\in\mathbb{R}^n$ and $y_0\in\mathbb{R}^N$, then a model with LinearScopes outputs:
\begin{equation}
    [y_0, y_1] = [f(x_0), F(x_0)\; x_1 + r(x_0)] \;,
\end{equation}
where $F(x_0)\in\mathbb{R}^{N\times n}$ is the \emph{filter} matrix and $r(x_0)\in\mathbb{R}^N$ is the \emph{residual}. A key idea proposed in DFV\cite{PNavarrete_2019a} is that we do not need to materialize the matrix $F(x_0)\in\mathbb{R}^{N\times n}$ to run the linear interpreter. The model with LinearScopes also avoids storage of activations in non--linear units because this information is used on--the--fly within LinearScopes (red lines in Figure \ref{fig:linearscopes}) and it is released afterwards.

Finally, our purpose will be to fix an input image $x_0$ and run tests with different probe inputs $x_1$ to get information from the linear interpreter.

\textbf{Residual and Columns:}
The procedure to calculate the residual $r(x_0)$ and columns of $F(x_0)$ from the linear interpreter follows the solution from DFV\cite{PNavarrete_2019a}. The residual is given by $y_1=r(x_0)$ when we use a probe batch $x_1=0$. Next, we can obtain a column $k$ from the filter matrix $F(x_0)$ as $y_1-r(x_0)$ when we use a probe batch $x_1=\delta_k$, where $\delta_k[k]=1$ and $\delta_k[i\neq k]=0$. This is an impulse response function according to signal processing theory\cite{JGProakis_2007a,SMallat_1998a}.

\textbf{Transposed System and Rows:}
To calculate $F^T(x_0)\cdot y_2$ for a given image in the output domain, $y_2\in\mathbb{R}^N$, we can use the vector calculus property for gradients of linear transformations: $\nabla_x(Ax+b) y = A^T y$. The same approach is used to implement (strided) transposed convolutions in deep learning frameworks\cite{parr2018matrix}, except that here our system is much bigger (possibly including transpose convolutions). Since deep learning frameworks provide automatic differentiation packages, it is simple and convenient to calculate:
\begin{equation}
    F^T(x_0)\cdot y_2 = \nabla_{x_1} y_1(x_1) \cdot y_2 \;. \label{eq:transpose}
\end{equation}
Finally, we can use the impulse response approach to obtain the rows of $F(x_0)$. This is, a row $k$ from the filter matrix $F(x_0)$ is given by $F^T(x_0)\cdot \delta_k$ when we use a probe image $y_2=\delta_k$, where $\delta_k[k]=1$ and $\delta_k[i\neq k]=0$.

Before moving forward, we emphasize that the transposed linear interpreter is different than the popular deconvolution method by Zeiler et.al\cite{DBLP:conf/eccv/ZeilerF14} because the deconvolution uses a non--linear output. More precisely, the procedure in \cite{DBLP:conf/eccv/ZeilerF14} describes how each layer must be transposed. The linear interpreter follows the same procedure for convolutional layers (linear) and max--pooling (our linear interpreter is equivalent to their approach), but for ReLU the approach in \cite{DBLP:conf/eccv/ZeilerF14} is to use an identical ReLU unit (non--linear). Instead, the linear interpreter will remember the activation of the unit in the forward pass (through gradients) and use the masking interpretation (linear).

\textbf{Singular Value Decomposition (SVD)}:
The dimension of inputs $x\in\mathbb{R}^n$ and outputs $y\in\mathbb{R}^N$ of a network can be different. Then the eigendecomposition of the filter matrix is given by its singular value decomposition (SVD). We propose Algorithm \ref{alg:power} to calculate the eigen--input/output for the largest singular value of $F(x_0)$, without materializing the matrix. We use an accelerated power method with momentum\cite{xu2018accelerated} adapted for SVD\cite{blum2016foundations}. Further eigen--inputs/outputs can be calculated in decreasing order of singular values by using a deflation method\cite{burden2010numerical,saad2011numerical}. For example, the second eigen--inputs/outputs and singular value is calculated by using Algorithm \ref{alg:power} on the \emph{deflated system} $F(x_0)+r(x_0)-\sigma_1 u_1 v^T_1$, and so forth.
\begin{algorithm}
    \centering
        \begin{algorithmic}[1]
            \REQUIRE Test image $x_0$.
            \REQUIRE Linear interpreter $y_1(x_1|x_0)$.
            \REQUIRE Residual $r(x_0)$.
            \REQUIRE Momentum $m$, number of steps $S$.
            \ENSURE $\sigma_{curr}$, $v_{curr}$, $u$.
            \STATE{
            $m \leftarrow 0$,
            $\sigma^2_{prev} \leftarrow 0$,
            $v_{prev} \leftarrow 0$,
            $v_{curr} \leftarrow \mathcal{N}(0,1)$
            }
            \FOR{$it = 1,\ldots,S$}
                \STATE{$u \leftarrow y_1(v_{curr}|x_0)-r(x_0)$}

                \STATE{$v_{next} \leftarrow F^T(x_0)\cdot u - m * v_{prev}$ \hfil use equation \eqref{eq:transpose}}

                \STATE{$\sigma^2_{curr} \leftarrow v^T_{curr} \cdot v_{next}$}

                \STATE{$v_{prev} \leftarrow v_{curr} / ||v_{next}||$}
                \STATE{$v_{curr} \leftarrow v_{next} / ||v_{next}||$}

                \STATE{$\sigma^2_{prev} \leftarrow \sigma^2_{curr}$}
            \ENDFOR
            \STATE{$u \leftarrow u / ||u||$}
        \end{algorithmic}
    \caption{SVD power method for a Linear Interpreter}
    \label{alg:power}
\end{algorithm}

\section{Experiments}
\begin{figure*}[ht!]
    \centering
    \includegraphics[width=\linewidth]{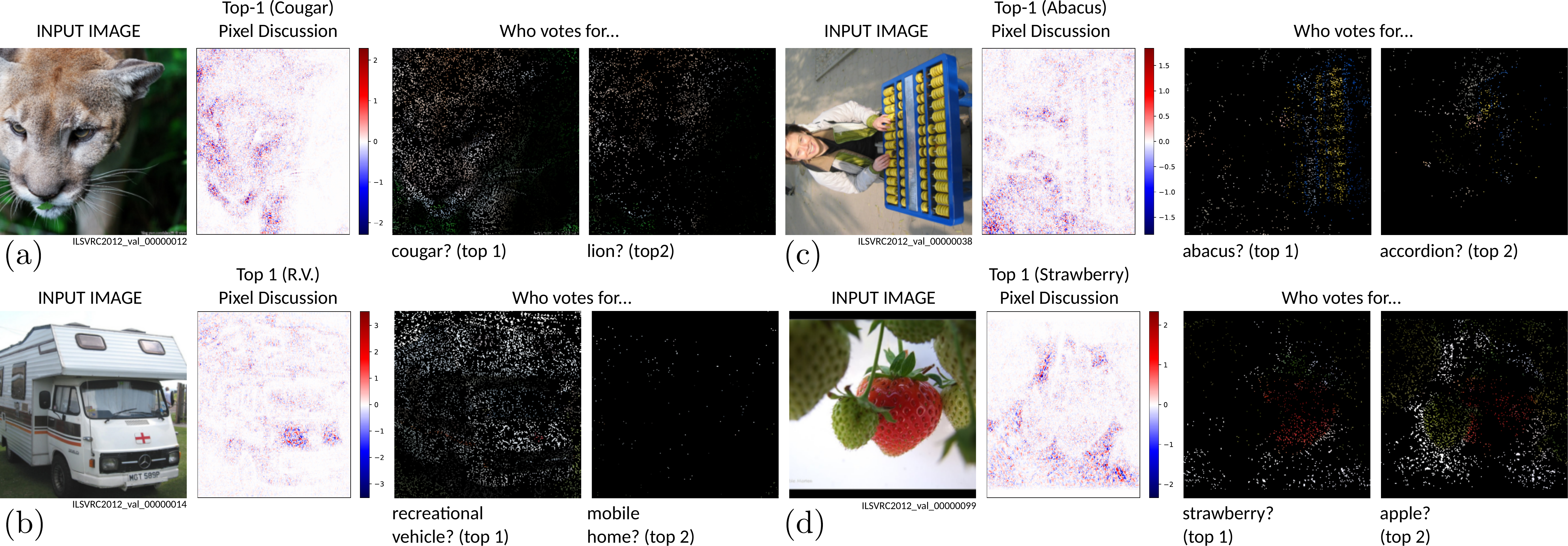}
    \caption{We back--project all the score contributions to input domain to show pixel--wise contributions, called \emph{pixel discussions} because pixels do not seem to agree on the scores. By comparing contributions among all scores, we make pixels vote independently and find that they finally focus on objects, with top--$2$ scores that show reasonable arguments for their votes.}
    \label{fig:classification_backprojections}
\end{figure*}
\begin{figure}[ht!]
    \centering
    \includegraphics[width=\linewidth]{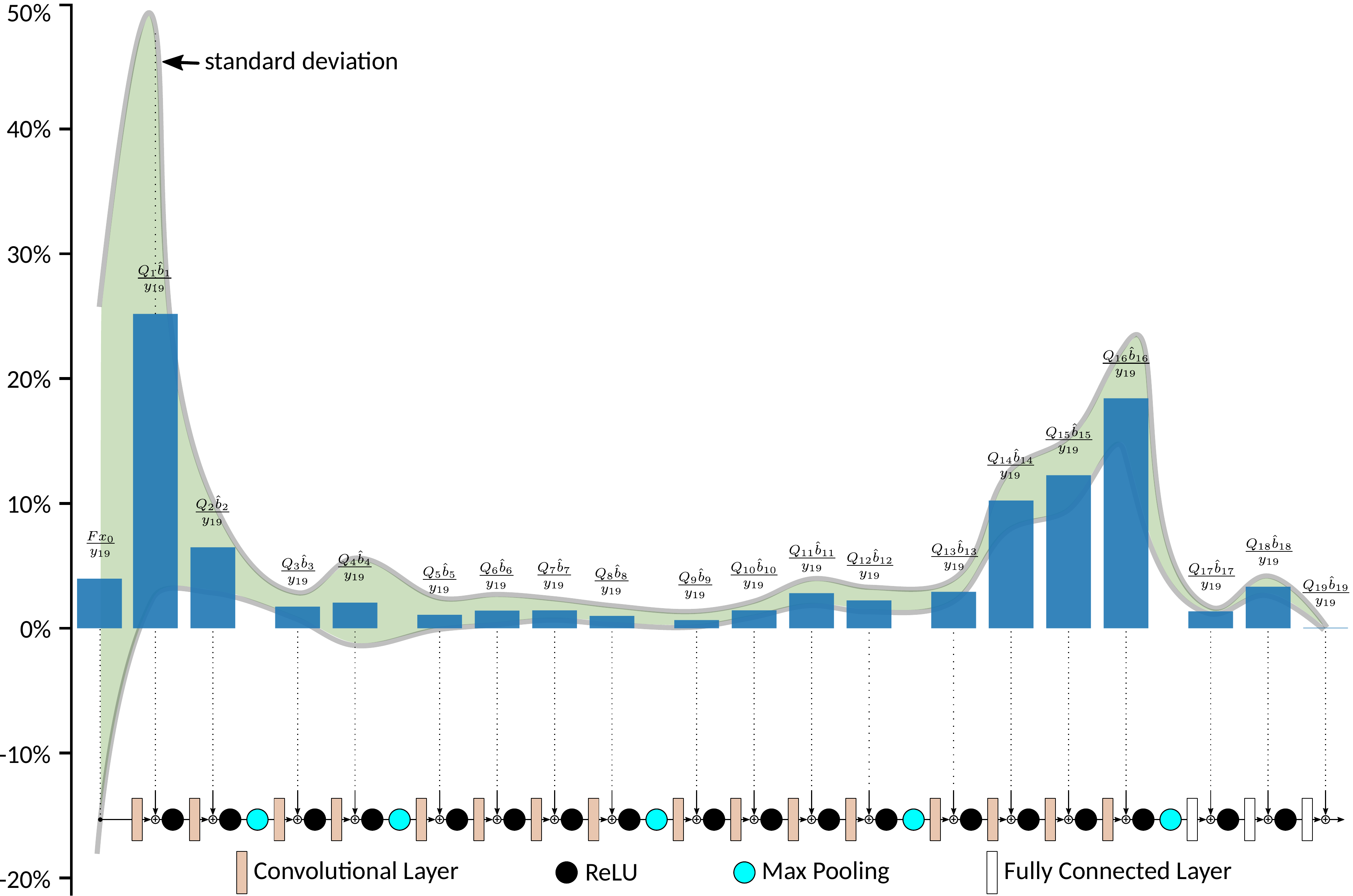}
    \caption{Layer--wise contributions to Top--$1$ scores for VGG--$19$ classifier\cite{Simonyan14c}, averaged over $100$ images from ImageNet--$1k$\cite{ILSVRC15} and normalized by the output score.\vspace*{-.1in}}
    \label{fig:classification_top1_distribution}
\end{figure}
\textbf{Case 1 -- Classification}: In this case a network takes images into scores (we do not include a softmax layer). If we look at a single score for a test image $x_0$ then $F(x_0)\in\mathbb{R}^{1\times n}$ is a single row image. Here, we are tempted to make a guess. We have seen evidence in DFV\cite{PNavarrete_2019a} that residuals are small. Then, if we want to maximize $F(x_0)x_0$ an ideal choice would be template--matching\cite{brunelli2009template, turin1960introduction}. This is, the network could try to construct a template image $F(x_0)$ that looks similar to $x_0$ for the correct label. In our experiments with various architectures we find that this is not the case. The image $F(x_0)$ does not look like a template and, most importantly, the residual $r(x_0)$ has the largest contribution to the scores, typically adding more than $80\%$ of the contribution as shown in Table \ref{tab:residuals}. This is a discouraging fact to conduct analysis since the residual of a score is a scalar that does not give more information than the score itself.
\begin{table}
    {\hfill
    \begin{tabular}{cccccc}
    \hline
    AlexNet & VGG--$19$ & ResNet--$152$ \\
    $78.5\%$ & $85.5\%$ & $81.1\%$ \\ \hline
    SqueezeNet 1.1 & DenseNet--161 & Inception v3 \\
    $84.3\%$ & $95.0\%$ & $91.6\%$ \\ \hline
    \end{tabular}
    \hfill}
    \caption{Average contributions of residuals for $100$ validation images from ImageNet--$1k$\cite{ILSVRC15}. The percentage increases for architectures with better benchmarks.} \label{tab:residuals}
\end{table}

But additional information can be obtained by using a theorem for sequential networks. For the sequential model:
\begin{equation}
    y_n = W_n x_{n-1} + b_n \quad\quad\text{and}\quad\quad x_n = h\left(y_n\right) \;,
\end{equation}
with parameters $b_n$ (biases) and sparse matrices $W_n$ (convolutions), we can get explicit formulas for the filter matrix and residual. This is:
\begin{thm}[from \cite{PNavarrete_2019a}] \label{thm:act_freeze}
Let $\hat{W}_n = A_n W_n$ and $\hat{b}_n = A_n b_n + c_n$. Where $A_n$, $c_n$ are the parameters of the linear interpreter of $h(y_n)$. Let
$Q_n = I$ and $Q_i = \prod_{k=i+1}^n \hat{W}_k$ for $i=1,\ldots,n$. The filter matrix and residual are:
\begin{equation}
    F = \prod_{k=1}^n \hat{W}_k\;, \quad\text{and}\quad r = \sum_{i=1}^n Q_i \hat{b}_i \;. \label{eq:sequential_parameters}
\end{equation}
\end{thm}

Let us grasp the meaning of this result. We will focus on networks with ReLU units so that $c_n=0$. First, the parameters with hat, $\hat{W}_n$ and $\hat{b}_n$, are the weights and biases of the network multiplied by masks. This already depends on the test image $x_0$. So, the formula for $F$ in \eqref{eq:sequential_parameters} basically represents the accumulated convolutions, masked by activations.

Next, matrices $Q_i$ represent the accumulated effect of convolutions, masked, from layer $i+1$ to $n$ (a forward projection). So finally, the formula for $r$ in \eqref{eq:sequential_parameters} gives us a \textbf{decomposition of the residual as layer--wise contributions of biases}, masked and forward projected into the scores.
\begin{figure*}[t]
    \centering
    \includegraphics[width=\linewidth]{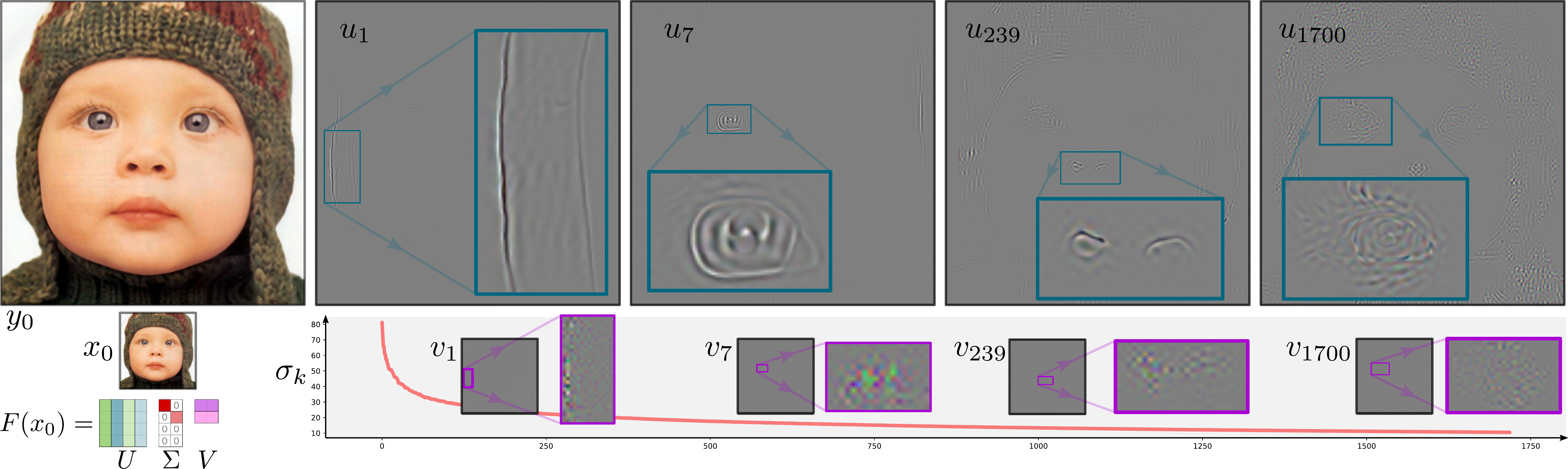}
    \caption{Results of the SVD of a linear interpreter applied on EDSR\cite{Lim_2017_CVPR_Workshops} $4\times$ super--resolution method.}
\label{fig:SR_svd}
\end{figure*}

In Figure \ref{fig:classification_top1_distribution} we show a histogram of the contributions for top--1 scores in a pre--trained VGG--$19$ network\cite{Simonyan14c}, averaged over $100$ validation images from ImageNet--$1k$\cite{ILSVRC15}. This includes a contribution of the input $F(x_0)x_0$ and the layer--wise contributions of the masked biases. We observe that most contributions come from the first two layers (with high variance) and the three layers before the fully--connected layers. For other variants of VGG we consistently observe two main contributions: one peak in early layers, and a second peak right before fully connected layers. But when the network is trained with batch normalization, the contributions move deeper in the network with one major contribution right before fully connected layers (see section \ref{app:classification}). Early contributions are based on local information as opposed to late contributions that use global information. This is reminiscent of results in \cite{montufar2014number} (section G) that use a similar linear mapping interpretation, discovering that hidden units learn to be invariant to more abstract translations at higher layers. In section \ref{app:classification} we also show how the contributions inside a network become random for images corrupted with adversarial noise using FGSM\cite{FGSM}, and final scores are exclusively due to the first few layers.

We can also perform a backward analysis by taking all the masked biases and back--project them from each layer to the input domain, adding them to $F(x_0)x_0$. We can perform this computation by considering subsystems from the input to an intermediate layer $k$ and use $F_k^T$ (using equation \eqref{eq:transpose}) on the masked biases $\hat{b}_k$. By summing all the back--projected contributions, we can see the pixel--wise contributions for each score. Examples are shown in Figure \ref{fig:classification_backprojections} for top--$1$ scores (more details in section \ref{app:classification}). We call these images \textbf{pixel discussions} because of the random behavior of pixels. They do not represent heatmaps because: first, highest values do not always focus on the objects; and second, positive values are followed by negative values in almost every pixel, as if pixels always digress with their neighbors on the contributions to the score. It should be noted that similar images are observed in LRP studies\cite{bach2015pixel,binder2016layer}.

Finally, we uncover clear information after we take each pixel contribution and compare it to the same pixel contributions for all other labels. In this way, \textbf{we make each pixel vote for a label}. In Figure \ref{fig:classification_backprojections} we mask the test image using the votes per pixel to observe what areas are more popular among pixels for a given label. The top--$1$ scores normally show the largest popularity and, most importantly, pixels clearly focus on objects. In Figure \ref{fig:classification_backprojections} (a) and (b), for example, pixels seem to discuss randomly on the face of a cougar and the lights of a vehicle, but when it comes to votes then distinctive features of the cougar appear as well as the whole vehicle. The votes for lion on \ref{fig:classification_backprojections} (a) show areas that could actually look more like a lion, so these pixels seem to have an argument. In Figure \ref{fig:classification_backprojections} (c) and (d), pixels discuss randomly in areas that do not contain the main object, but after voting they do focus on the objects. Figure \ref{fig:classification_backprojections} (d) is interesting because the votes for strawberry show the red shape of a strawberry, and the votes for apple do show green and red shapes that resemble a couple of apples.

\begin{figure}
    \centering
    \includegraphics[width=.9\linewidth]{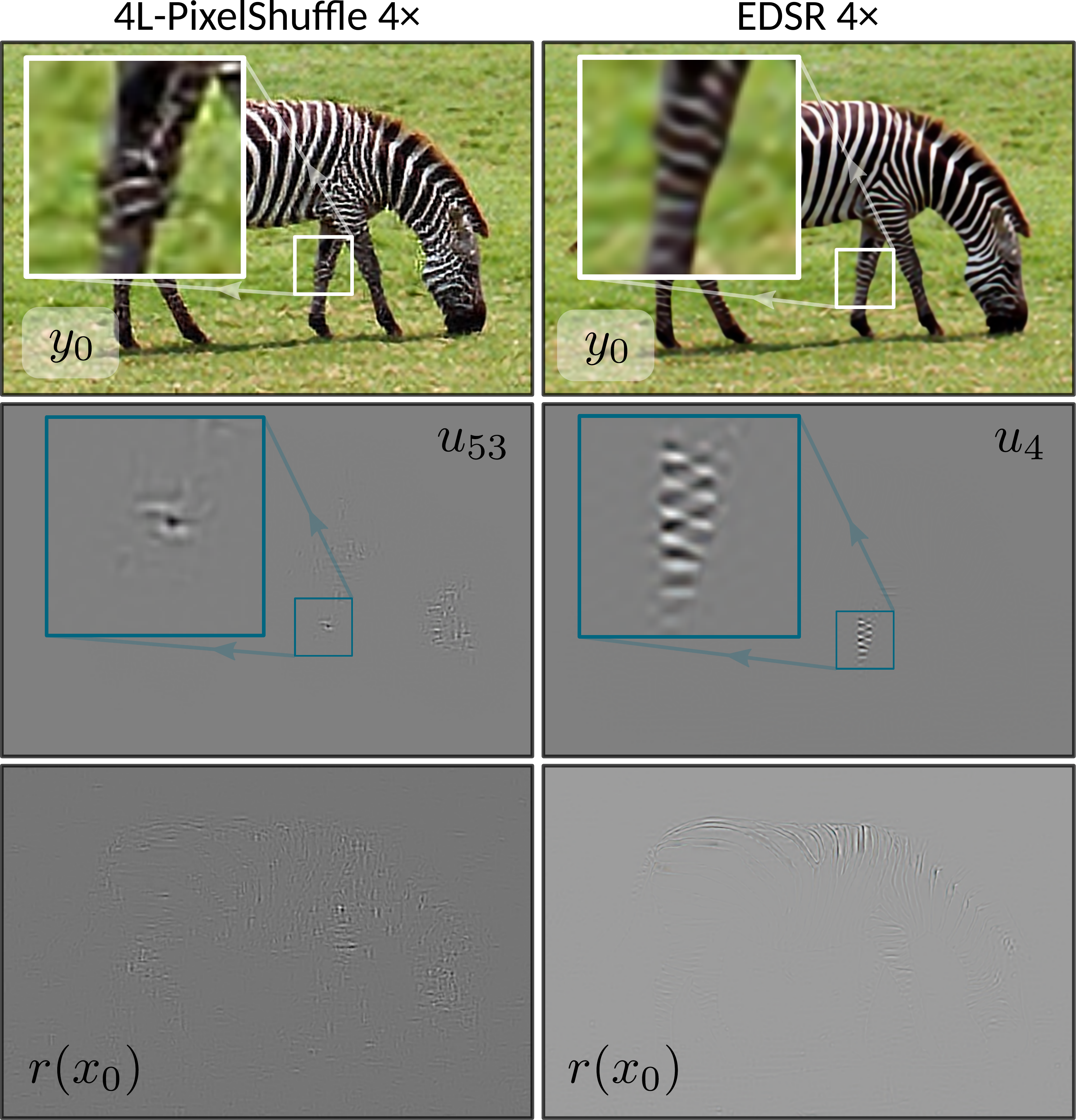}
    \caption{The SVD of SR models show how better models (EDSR) capture higher--level features from images.\vspace*{-.1in}}
\label{fig:SR_compare}
\end{figure}
\textbf{Case 2 -- Super--Resolution (SR)}:
In this case a network takes small images into large images. The filter matrix $F(x_0)\in\mathbb{R}^{N\times n}$, with $N>n$, has a tall rectangular shape. The linear interpreter analysis was originally used in DFV\cite{PNavarrete_2019a} to study this problem. In \cite{PNavarrete_2019a} only projective filter coefficients were obtained (columns of the filter matrix). We show results with receptive filter coefficients in section \ref{app:sr}, which are more closely related to the traditional concept of convolutional filters. In addition, we can now efficiently calculate all the rows and columns for a given image, using very big models such as EDSR\cite{Lim_2017_CVPR_Workshops} (see demonstrations in section \ref{app:links}).

Figure \ref{fig:SR_svd} shows examples of the eigen--inputs/outputs and singular values of EDSR\cite{Lim_2017_CVPR_Workshops} $4\times$ upscaler. Before we interpret these results it is convenient to remember a simple reference. A classic upscaler uses linear--space--invariant (LSI) filters\cite{JGProakis_2007a,SMallat_1998a} whose eigen--inputs/outputs are harmonic functions (e.g. some type of DCT). So, our reference from classic upscaling are basis that cover all the image using different frequencies. The information in Figure \ref{fig:SR_svd} reveals a very different approach followed by convolutional networks. First, we observe oscillations of high frequencies in the eigen--inputs. These are similar to high frequency stimulus used in psychovisual experiments of contrast sensitivity function, where subjects are required to view sequential simple stimuli, like sine--wave gratings or Gabor patches\cite{watanabe1968spatial}. The response of the network to these stimuli are clear pieces of images (e.g. an eye, a corner, a nose, etc.), smooth and localized in space for high singular values, and extending in space with higher frequency components for lower singular values. So the network reacts to stimulus similar to Gabor wavelets by triggering image objects. The response is similar to the receptive fields of simple cells in mammalian primary visual cortex that can be characterized as being spatially localized, oriented and bandpass, comparable to wavelet basis\cite{olshausen1996emergence,SMallat_1998a}. Compared to EigenFaces obtained by PCA decompositions\cite{Sirovich:87}, we observe a similar pattern of low to high frequency oscillations as the eigen/singular--values reduce. But EigenFaces are not localized like the CNN eigen--decomposition in Figure \ref{fig:SR_svd}.

Finally, in Figure \ref{fig:SR_compare} we show how an SVD analysis helps to evaluate models. A $4$--layer PixelShuffle model commonly used in deep--learning tutorials is compared to EDSR model. The image quality of EDSR is clearly better. We observe that residuals are small for SR models. For EDSR the residual is more focused on the back and neck of the zebra, whereas the residual in PixelShuffle is spread all over the image. In the eigen--outputs we see that EDSR focuses in features that are visible parts of the zebra. The eigen--output where the PixelShuffle model focuses on the same area (back leg), does not show clear local features of the zebra. We can conclude that better models are able to capture and focus on high--level objects in an image.

\textbf{Case 3 -- Image--to--Image Translation (I2I)}:
\begin{figure}[t]
    \centering
    \includegraphics[width=.9\linewidth]{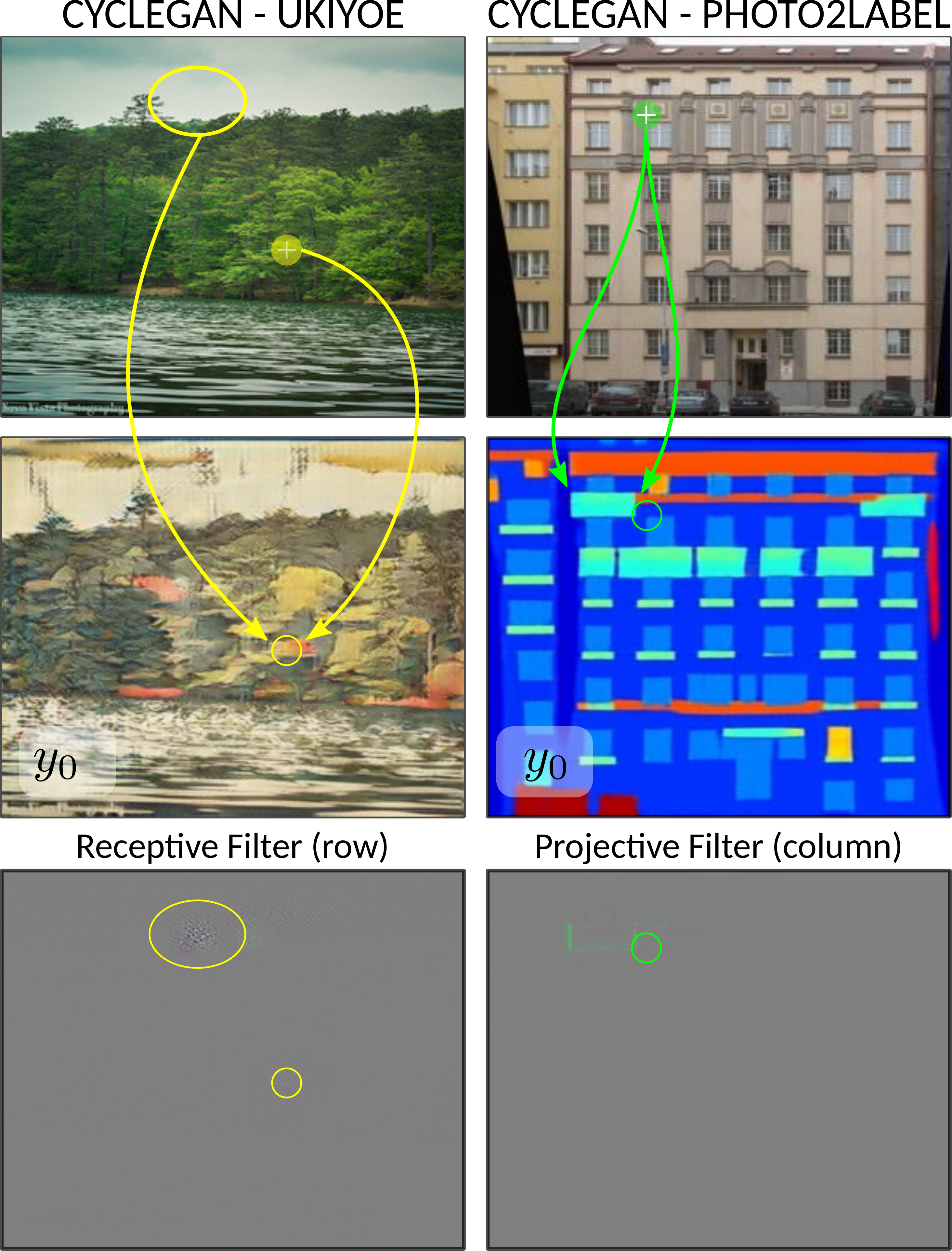}
    \caption{Receptive and Projective filters of the linear interpreter for CycleGAN\cite{CycleGAN2017} Ukiyoe and Facades. An off--diagonals (yellow ellipsis) is used in Ukiyoe to help generating textures. A single pixel helps to create a template window box in Facades.\vspace*{-.1in}}
\label{fig:I2I_filters}
\end{figure}
We end our tour with a network that does not change the size of images. The filter matrix $F(x_0)\in\mathbb{R}^{N\times n}$, with $N=n$, is square. Here, we choose to test different pre--trained models of the popular CycleGAN architecture\cite{CycleGAN2017}. This architecture uses instance--normalization layers that are known to improve visual effects by using global information (means and variances of network features). For this, we use the linear interpreter shown in Figure \ref{fig:linearscopes}.

In Figure \ref{fig:I2I_filters} we show projective and receptive filter coefficients for two I2I tasks: image--to--painting (similar to style transfer) and photo--to--facade (similar to segmentation). On one hand, compared to SR, the I2I tasks show some similarities. In most areas of an image we observe localized filter coefficients (see demonstrations in section \ref{app:links}) which means that the filter matrix is sparse and concentrated around the diagonal, similar to SR. But on the other hand, the receptive/projective fields are larger in CycleGAN and the most distinctive feature is the appearance of strong off--diagonals. Figure \ref{fig:I2I_filters} shows how in photo--to--painting the receptive filter uses information localized to a particular output location (small circle) and adds significant information from an area in the upper part of the image (the ellipsis). We observe that for a single image, CycleGAN consistently uses the same area (e.g. the ellipsis in Figure \ref{fig:I2I_filters}) to pass information to all other pixels in the image. This \textbf{copy--move strategy} seems to give the ability to create a consistent texture all over the image, taken from a fixed place and combined with the local pixels.

In the photo--to--facade task, besides the appearance of strong off-diagonals, we observe how single pixels are directed to specific segments of the output. By this means, CycleGAN \textbf{creates templates} (e.g. window boxes) that are usually triggered by pixels in corner or edges as shown in Figure \ref{fig:I2I_filters}. Also, for this case, the receptive filter coefficients can sometimes extend to the whole image (see demonstration in section \ref{app:links}). This behavior is only possible due to instance normalization layers carrying global information of the image. In SR tasks, usually trained over relatively small patches (e.g. $48\times 48$ in small resolution) a network cannot learn such strategies. Pretrained models of CycleGAN used whole images ($256\times 256$) for training.

Results of SVD decomposition for CycleGAN are included in section \ref{app:I2I}. Here, the eigen--inputs/outputs show similar patters to SR but the stimuli and responses in the output cover much larger areas and show several objects in the eigen--outputs as opposed to single objects observed in SR. This is likely caused by off--diagonal patterns.

\section{Discussion}
\label{sec:discussion}
LRP\cite{bach2015pixel} introduces the concept of relevance of each pixel in the classification scores.  If we use our layer--wise contributions to redefine LRP relevances we could force our analysis to fit into the LRP framework. Our contributions are significant because of the novel interpretation, revealing an explicit contribution of biases to the final scores that was previously unknown. At pixel level, LRP has been used to study the influence of input pixels to the final scores in order of pixel--wise relevances\cite{binder2016layer}. On the other hand, pixel--discussions can be used independent of the scores to obtain the vote of each pixel. Besides this difference, further investigation is necessary to better understand the relationship between pixel--discussions and other heatmap/saliency visualizations.

DTD\cite{montavon2017explaining} uses layer--wise Taylor expansions and modifies the root points to obtain heatmaps that are consistent (conservative and positive). In our analysis we do not control the backprojections leading to pixel--discussions and as a result we find that they do not work as heatmaps but as independent votes. The targets and results of interpretability compared to DTD are therefore different, but further investigation is necessary to better understand this relationship.

Finally, our approach in this paper relies on the human understanding of linear systems. Therefore, the effect of visualization results on human understanding is not direct. Future research is necessary to understand whether humans can predict model failures better, as proposed in \cite{doshi2017roadmap}, with or without access to LinearScope visualizations.

\section{Conclusions}
We introduced a hooking layer, called a \textbf{\mbox{LinearScope}}, that allows to run a network and its linear interpreter in parallel. By efficiently running a linear interpreter, it allows more powerful analysis of CNN architectures. We explored three applications to emphasize the generality of this approach and how it can be used to interpret the different ways in which convolutional networks adapt to the problems.

\section{Acknowledgements}
The authors would like to thank Xiaomin Zhang for constructive criticism of the manuscript.

\section{Appendix}
\def\thesubsection{\arabic{section}.\Alph{subsection}}
\emph{
We provide the following additional information:
\begin{itemize}
    \item Classification:
    \begin{itemize}
        \item Explanation of Forward/Back--Projections;
        \item Residual contributions for more architectures;
        \item Contribution histograms for more networks;
        \item Pixel votes for more images;
        \item What happens after an adversarial attack?
    \end{itemize}
    \item Super--Resolution (SR):
    \begin{itemize}
        \item Projective/receptive filters;
        \item More eigen--inputs/outputs.
    \end{itemize}
    \item Image--to--Image Translation (I2I):
    \begin{itemize}
        \item Projective/receptive filters;
        \item Eigen--inputs/outputs.
    \end{itemize}
    \item Demonstrations
    \begin{itemize}
        \item Video material;
        \item Interactive material.
    \end{itemize}
\end{itemize}
}
\subsection{Classification}
\label{app:classification}
\textbf{Explanation of Forward/Back--Projections}:

In our analysis of linear interpreters for classifiers we use a theorem that is essential to understand how do we decompose the contributions of the network to the output scores. Roughly speaking, the theorem says that:
    \begin{quote}
        \emph{In a sequential network there are explicit expressions for $F$ and $r$ in the linear interpreter $y=Fx+r$. The filter matrix $F$ is given by the forward--projection of the input to the output score. And $r$ is given by the sum of all forward--projected masked--biases from each layer to the output score.}
    \end{quote}
By \emph{projection} we mean the progressive application of the linear transformation for each layer. \textbf{Forward projection} means that we apply the linear transformations of a given layer, and then the transformation of the next layer, and so forth. \textbf{Backward projection} means that we apply the transposed linear transformation of a given layer, and then the same in the previous layer, and so forth. Finally, \textbf{masked--biases} are the bias parameters of the network (scalars) multiplied by activation masks (images of ones and zeros for ReLU). Then, the masked--biases, denoted by $\hat{b}$, are images in the network's feature domain at the layer, that can be forward or back projected through the network.

The proof of the theorem is straightforward using inductive arguments. So here we prefer to follow a more didactic approach. Namely, we will unfold the formula for the linear interpreter and see how the expressions for filter matrix and residual decomposition appear.

We start with a sequential convolutional network model:
\begin{equation}
    y_n = W_n x_{n-1} + b_n \quad\quad\text{and}\quad\quad x_n = h\left(y_n\right) \;,
\end{equation}
with parameters $b_n$ (biases) and sparse matrices $W_n$ (convolutions, including strided and transposed).

Let $\hat{W}_n = A_n W_n$ and $\hat{b}_n = A_n b_n + c_n$. Where $A_n$, $c_n$ are the parameters of the linear interpreter for $h(y_n)$. Then we have:
\begin{align}
    x_n & = h\left( W_n h\left(W_{n-1} x_{n-2} + b_{n-1}\right) + b_n \right) \\
        & = \hat{W}_n \left(\hat{W}_{n-1} x_{n-2} + \underbrace{A_{n-1}b_{n-1} + c_{n-1}}_{\hat{b}_{n-1}}\right) + \underbrace{A_n b_n + c_n}_{\hat{b}_n}\\
        & = \hat{W}_n\hat{W}_{n-1} \underbrace{x_{n-2}}_{\hat{W}_{n-2} x_{n-3} + \hat{b}_{n-2}} + \hat{W}_n \hat{b}_{n-1} + \hat{b}_n \\
        & = \prod_{k=1}^n \hat{W}_k x_0 + \prod_{k=2}^n \hat{W}_k \hat{b}_1 + \prod_{k=3}^n \hat{W}_k \hat{b}_2 + \cdots + \hat{b}_n \;.
\end{align}
Now we can define:
\begin{equation}
    Q_n = I \;, \quad Q_i = \prod_{k=i+1}^n \hat{W}_k, \quad\text{for } i=1,\ldots,n \;.
\end{equation}
and we get:
\begin{equation}
    \boxed{x_n = \underbrace{Q_0}_{F} x_0 + \underbrace{Q_1 \hat{b}_1 + Q_2 \hat{b}_2 + \cdots + \hat{b}_n}_{r}} \;.
\end{equation}
The filter matrix and residual at layer $n$ are then given by:
\begin{equation}
    F = \prod_{k=1}^n \hat{W}_k\;, \quad\text{and}\quad r = \sum_{i=1}^n Q_i \hat{b}_i \;.
\end{equation}
This expression follows the so--called \emph{conservation property} of LRP\cite{bach2015pixel} because the final score (the output of the network) is written as a sum of layer--wise contributions. Nevertheless, the nature of these contributions here has a different meaning, not as relevances but as forward--projected masked--biases.

\textbf{Matrices $Q$ represent forward--projections}, since they progressively apply linear transformations towards the output. Similarly, we can define:
\begin{equation}
    P_0 = I \;, \quad P_i = \prod_{k=1}^i \hat{W}^T_k, \quad\text{for } i=1,\ldots,n \;.
\end{equation}
Here, \textbf{matrices $P$ represent back--projections}, since they progressively apply transposed linear transformations towards the input.

We can use this definition to give an explicit expression for the \textbf{Pixel Discussion} (PD) images displayed in the main text. This is
\begin{equation}
    \boxed{\text{PD} \propto P_0 F^T(x_0) + P_1 \hat{b}_1 + P_2 \hat{b}_2 + \cdots + P_n \hat{b}_n} \;,
\end{equation}
and PD is normalized so that the sum of all of its pixels gives us the output score. Then, each pixel value in PD gives a pixel--wise contribution to the final score.

\bigskip
\textbf{Residual contributions for more architectures}:
\begin{table}
    {\hfill
    \begin{tabular}{cp{.01cm}cp{.01cm}c}
    \cline{1-1}\cline{3-3}\cline{5-5}
    AlexNet & & SqueezeNet $1.0$ & & VGG--$11$ \\
    $\stackrel{\displaystyle78.5\%}{\scriptscriptstyle\pm15.8}$ & & $\stackrel{\displaystyle80.7\%}{\scriptscriptstyle\pm11.5}$ & & $\stackrel{\displaystyle82.2\%}{\scriptscriptstyle\pm14.1}$ \\ \cline{1-1}
    ResNet--$18$ & & SqueezeNet $1.1$ & & VGG--$13$ \\
    $\stackrel{\displaystyle80.5\%}{\scriptscriptstyle\pm13.9}$ & & $\stackrel{\displaystyle84.3\%}{\scriptscriptstyle\pm11.0}$ & & $\stackrel{\displaystyle82.11\%}{\scriptscriptstyle\pm13.3}$ \\ \cline{3-3}
    ResNet--$34$ & & DenseNet--$121$ & & VGG--$16$ \\
    $\stackrel{\displaystyle83.7\%}{\scriptscriptstyle\pm12.9}$ & & $\stackrel{\displaystyle94.6\%}{\scriptscriptstyle\pm4.5}$ & & $\stackrel{\displaystyle84.2\%}{\scriptscriptstyle\pm12.0}$ \\
    ResNet--$50$ & & DenseNet--$161$ & & VGG--$19$ \\
    $\stackrel{\displaystyle82.0\%}{\scriptscriptstyle\pm16.4}$ & & $\stackrel{\displaystyle95.0\%}{\scriptscriptstyle\pm4.0}$ & & $\stackrel{\displaystyle85.5\%}{\scriptscriptstyle\pm10.9}$ \\ \cline{5-5}
    ResNet--$101$ & & DenseNet--$169$ & & VGG--$11$--BN \\
    $\stackrel{\displaystyle80.2\%}{\scriptscriptstyle\pm13.6}$ & & $\stackrel{\displaystyle94.4\%}{\scriptscriptstyle\pm4.4}$ & & $\stackrel{\displaystyle84.5\%}{\scriptscriptstyle\pm12.6}$ \\
    ResNet--$152$ & & DenseNet--$201$ & & VGG--$13$--BN \\
    $\stackrel{\displaystyle81.1\%}{\scriptscriptstyle\pm16.9}$ & & $\stackrel{\displaystyle94.0\%}{\scriptscriptstyle\pm4.8}$ & & $\stackrel{\displaystyle85.1\%}{\scriptscriptstyle\pm12.5}$ \\ \cline{1-1} \cline{3-3}
    & & Inception v3 & & VGG--$16$--BN \\
    & & $\stackrel{\displaystyle91.6\%}{\scriptscriptstyle\pm8.0}$ & & $\stackrel{\displaystyle86.2\%}{\scriptscriptstyle\pm12.9}$ \\ \cline{3-3}
    & & & & VGG--$19$--BN \\
    & & & & $\stackrel{\displaystyle85.8\%}{\scriptscriptstyle\pm13.6}$ \\ \cline{5-5}
    \end{tabular}
    \caption{Average contributions of residuals to classification scores for $100$ validation images from ImageNet--$1k$\cite{ILSVRC15}. Numbers below percentage represent standard deviation. DenseNet and Inception architectures show highest contributions of the residual, with smaller standard deviation.} \label{tab:residuals_all}
    \hfill}
\end{table}

In Table \ref{tab:residuals_all} we show the average contribution of residual to classification scores for a more complete list of architectures, including standard deviation values. In VGG and SqueezeNet we observe that the residual contribution increases for larger networks (with better benchmarks) but this pattern does not repeat for other architectures. Standard deviations are smaller for larger contributions of the residual, indicating that these architectures are consistently using the residual to improve their classification scores.

\bigskip
\textbf{Contribution histograms for more networks}:
\begin{figure*}
    \centering
    \includegraphics[width=\linewidth]{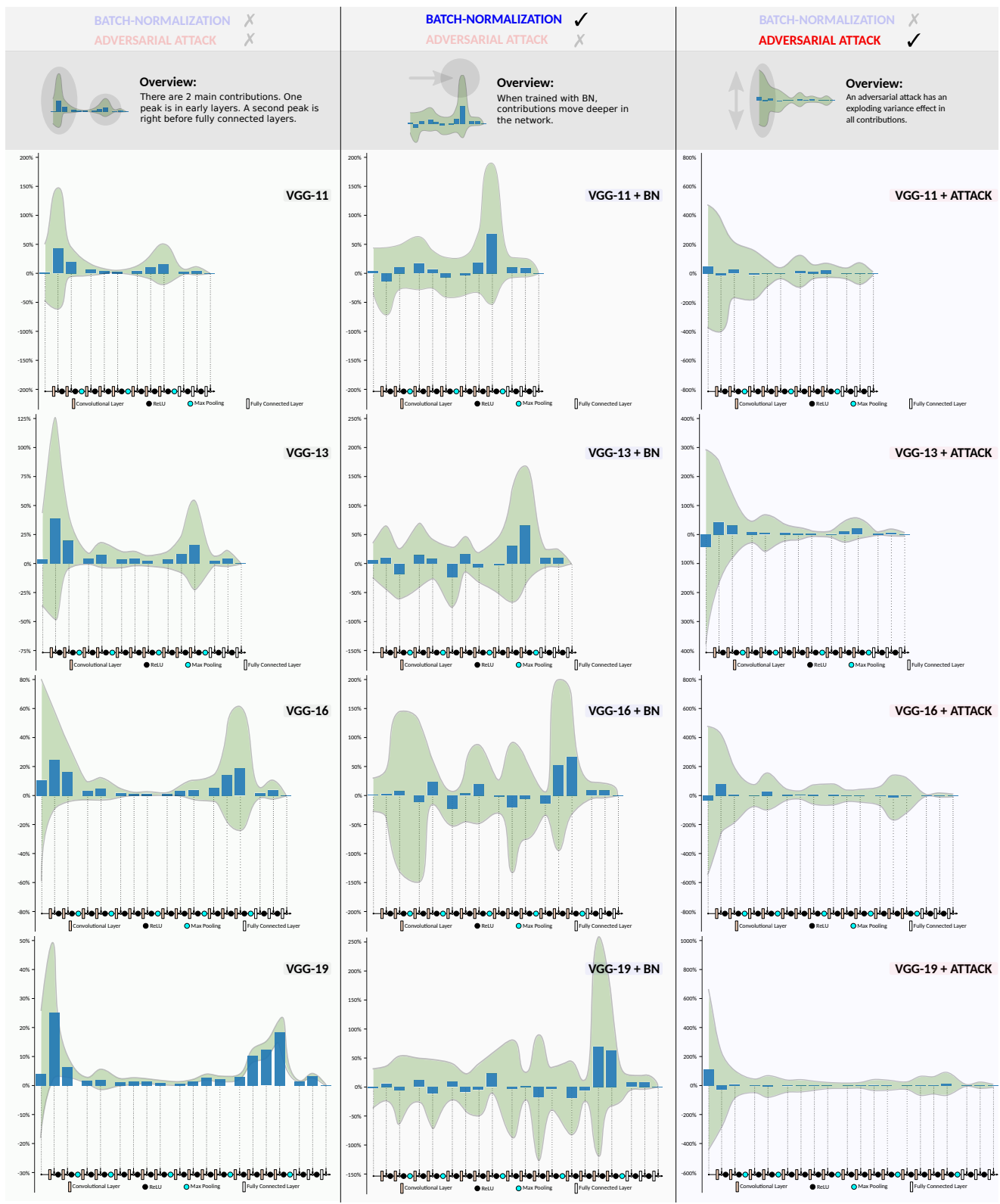}
    \caption{Layer--wise contributions to Top--$1$ scores for pre--trained VGG classifiers, averaged over $100$ images from ImageNet--$1k$. Standard deviation shown as shaded area. The first column shows models trained with original images and without batch--normalization. The second column uses original images and models with batch--normalization. The third column considers the same group of $100$ images with adversarial attack added by using FGSM\cite{FGSM}.}
\label{fig:vgg_histograms}
\end{figure*}

In Figure \ref{fig:vgg_histograms} we show histograms of the layer--wise contributions to top--$1$ scores for a series of VGG network architectures. We use pre--trained models trained with and without batch--normalization\footnote{Classifier models downloaded from \url{https://pytorch.org/docs/stable/torchvision/models.html}}. We observe in most cases that the contribution of the input, $F(x_0)x_0$, does not account for the largest part of the final score. So it is necessary to use the layer--wise decomposition of the residual to really see where do the contributions come from.

When trained without batch--normalization, we consistently see two major contributions. One in early layers of the network (before the first pooling layer). And the second major contribution comes from much deeper in the network, just before the fully connected layers. This pattern clearly changes in networks trained with batch--normalization. In this case the contributions move inside the network, with major contributions just before fully connected layers.

\bigskip
\textbf{Pixel votes for more images}:
\begin{figure*}
    \centering
    \includegraphics[width=\linewidth]{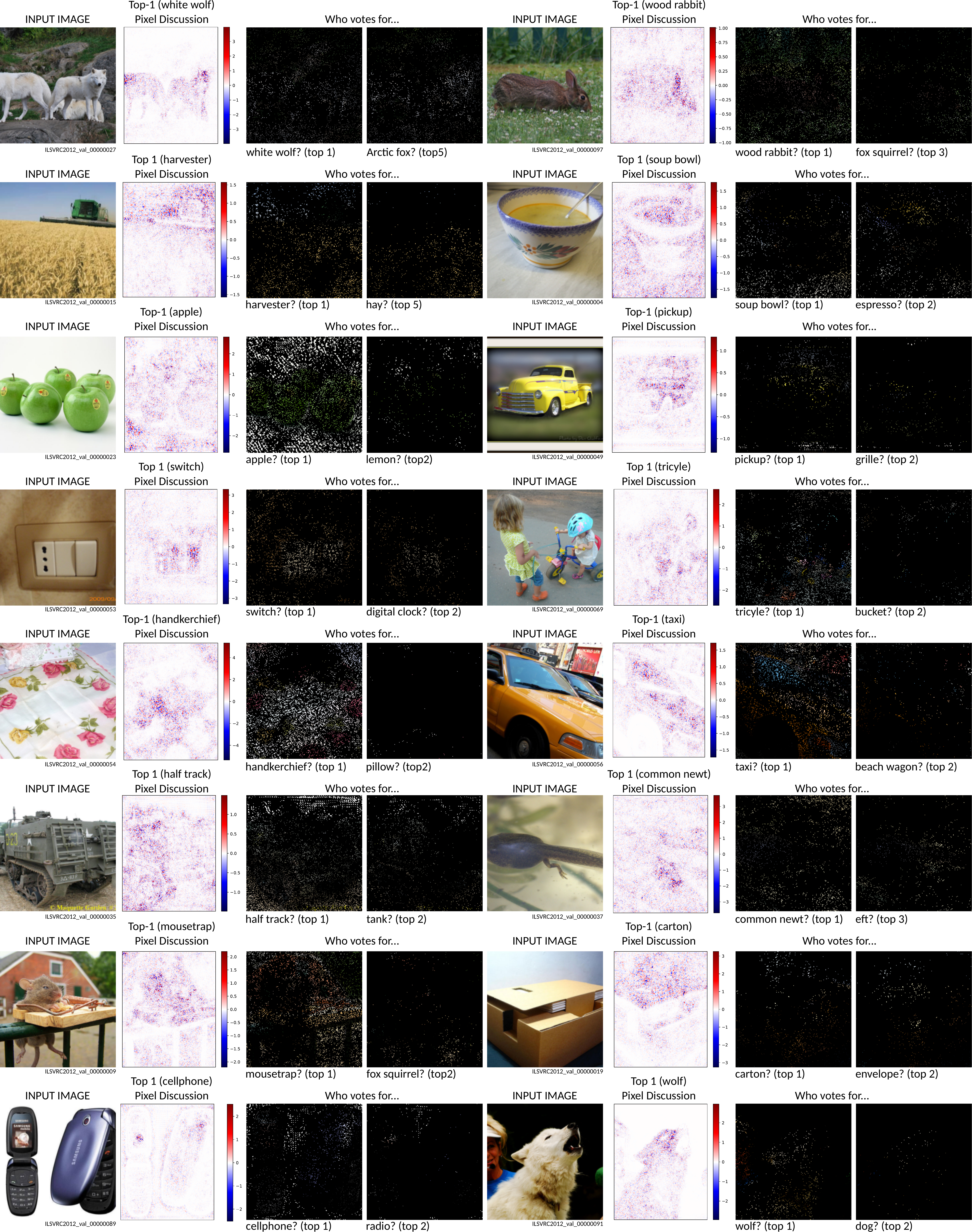}
    \caption{Pixel--discussions are back--projections of output scores to input domain that show pixel--wise contributions to the scores. By comparing contributions among all scores, we make pixels vote independently and find that they focus on objects.}
\label{fig:backprojection_original}
\end{figure*}
\begin{figure*}
    \centering
    \includegraphics[width=\linewidth]{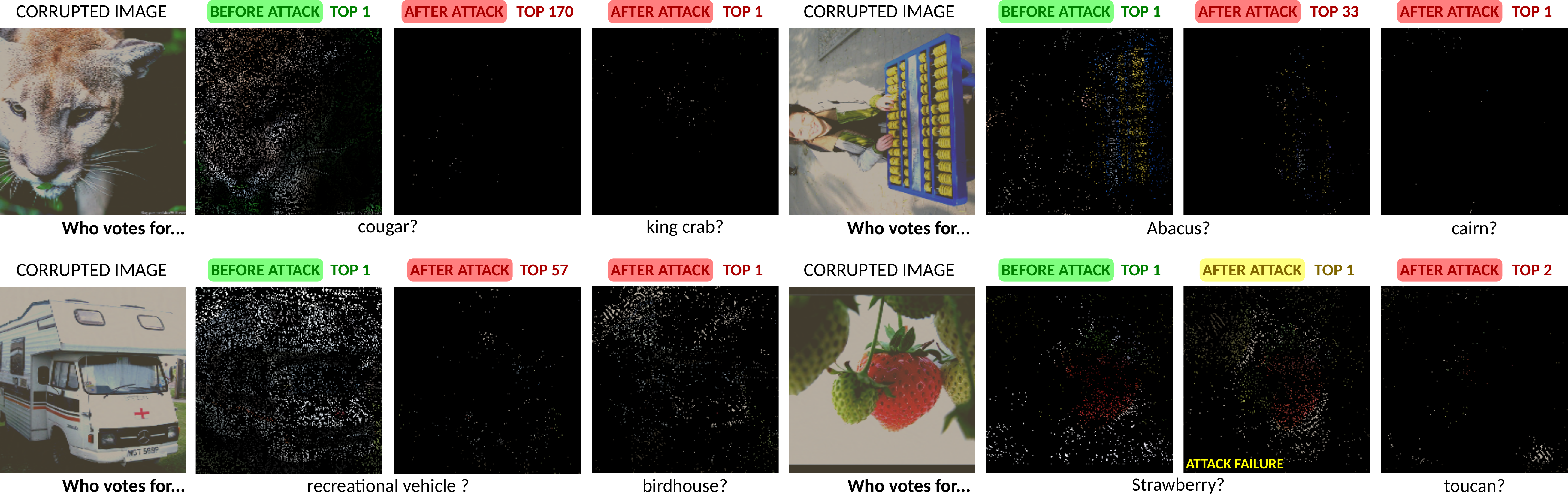}
    \caption{Effect of an adversarial attack using FGSM\cite{FGSM} on the votes of pixels. When an attack succeeds the pixels clearly stop to vote for the right label. In most cases the pixels do not seem to vote much for the new top--$1$ label, suggesting that the attack is spreading the opinion of pixels throughout all the $1,000$ classes.}
\label{fig:backprojection_attack}
\end{figure*}

In Figure \ref{fig:backprojection_original} we show more examples of pixel discussions and pixel votes. Here, we observe more evidence that pixel dicussions (PDs) are not conclusive about the network's decision. Sometimes, pixels seem to discuss strongly on an object (e.g. wolf, pickup car, taxi, etc.) but in other cases the discussion takes place outside the main object (e.g. harvester, soup bowl). After we compare the discussions over all labels we can see what is the overall pixel--wise outcome of the discussion. These so--called \emph{pixel votes} focus on the main objects and show clear preferences for the top score in the output of the network. We observe how ``harvest'' and ``hay'' labels have pixels focused on areas with hay; an ``espresso'' label have pixels looking at the liquid in a bowl; a ``digital clock'' label have pixels on a square--shape plug that looks like a digital clock; etc.

\bigskip
\textbf{What happens after an adversarial attack?}

We have observed clear patterns in the contribution histograms and pixel votes that show the layers where networks make decisions and the pixel--wise preferences for each label. To explore these patterns further, now we consider the effect of an adversarial attack on the network. Namely, we consider a \emph{Fast Gradient Sign Method}\cite{FGSM} (FGSM)\footnote{Attack implemented by using code from \url{https://github.com/baidu/AdvBox}.} that introduces noise in input images, making them look brighter but keeping the content visible to human eyes. In Figure \ref{fig:backprojection_attack} we observe how the attack changes the decision of the network without displaying visible content of the new top labels in the corrupted images. Here, we observe a strong change in the pattern of pixel votes. In one case (strawberry) where the attack fails, we still see the pixels voting more for top--$1$ label. When the attack succeeds, pixels stop to vote for the right label but they also do not vote much for the new top labels. It seems then that the effect of the attack is to spread the votes of pixels throughout all $1,000$ classes. This hypothesis is consistent with the effect on the contribution histograms. In Figure \ref{fig:vgg_histograms} we observe that the attack has a strong effect on the variance of the contributions per layer. So for each image we get a different histogram, with strong positive and negative contributions. \textbf{The network does not behave normal with images corrupted with adversarial attacks}. The layers do not contribute in the same way and pixel votes do not show strong agreements.

\subsection{Super--Resolution (SR)}
\label{app:sr}
\begin{figure*}
    \centering
    \framebox{\includegraphics[width=.3\linewidth]{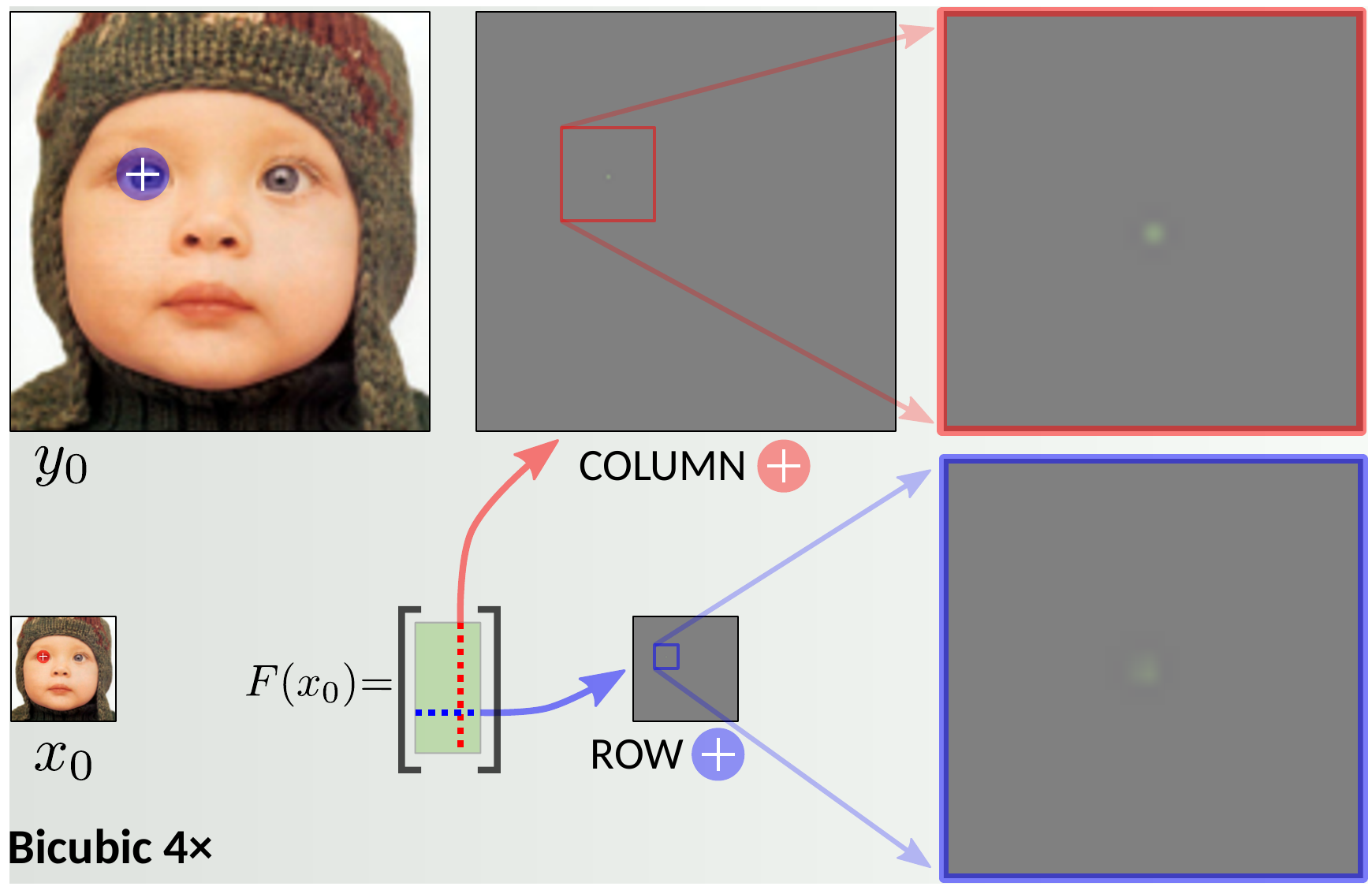}}
    \framebox{\includegraphics[width=.3\linewidth]{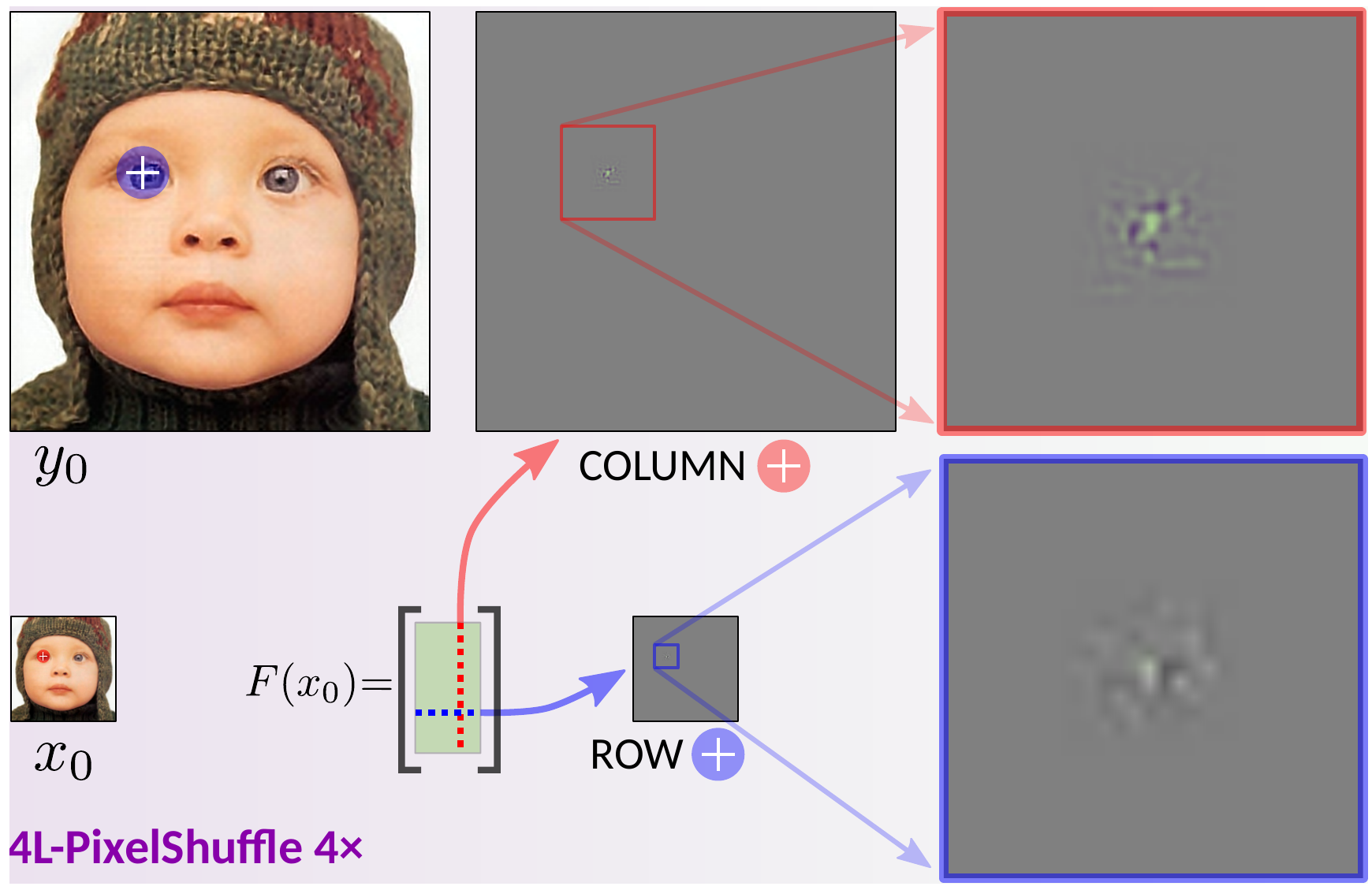}}
    \framebox{\includegraphics[width=.3\linewidth]{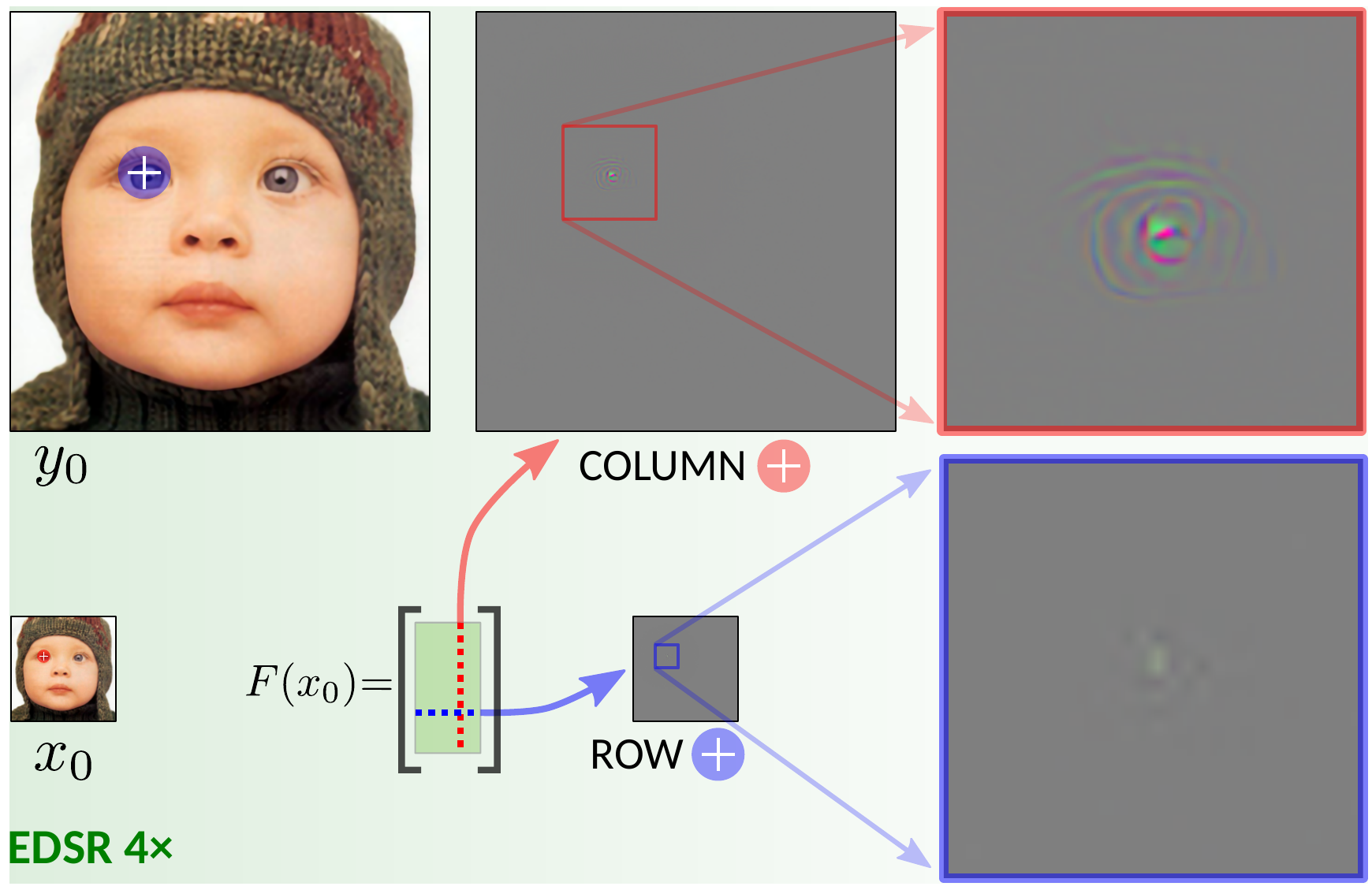}}
    \caption{Screenshots of live demonstration showing the projective and receptive filters (columns and rows) for the linear interpreter of upscaler systems. A bicubic upscaler does not adapt to the image and keeps the filter coefficients unchanged. The $4$--layer PixelShuffle model adapts to the image, changing on edges and textures, but does not clearly follow the geometry. The EDSR\cite{Lim_2017_CVPR_Workshops} model adapts to the image and reveals the geometry of high--level features (e.g. eyes, textures, nose).}
\label{fig:FilterMatrix_SR}
\end{figure*}
\begin{figure*}
    \centering
    \includegraphics[width=.99\linewidth]{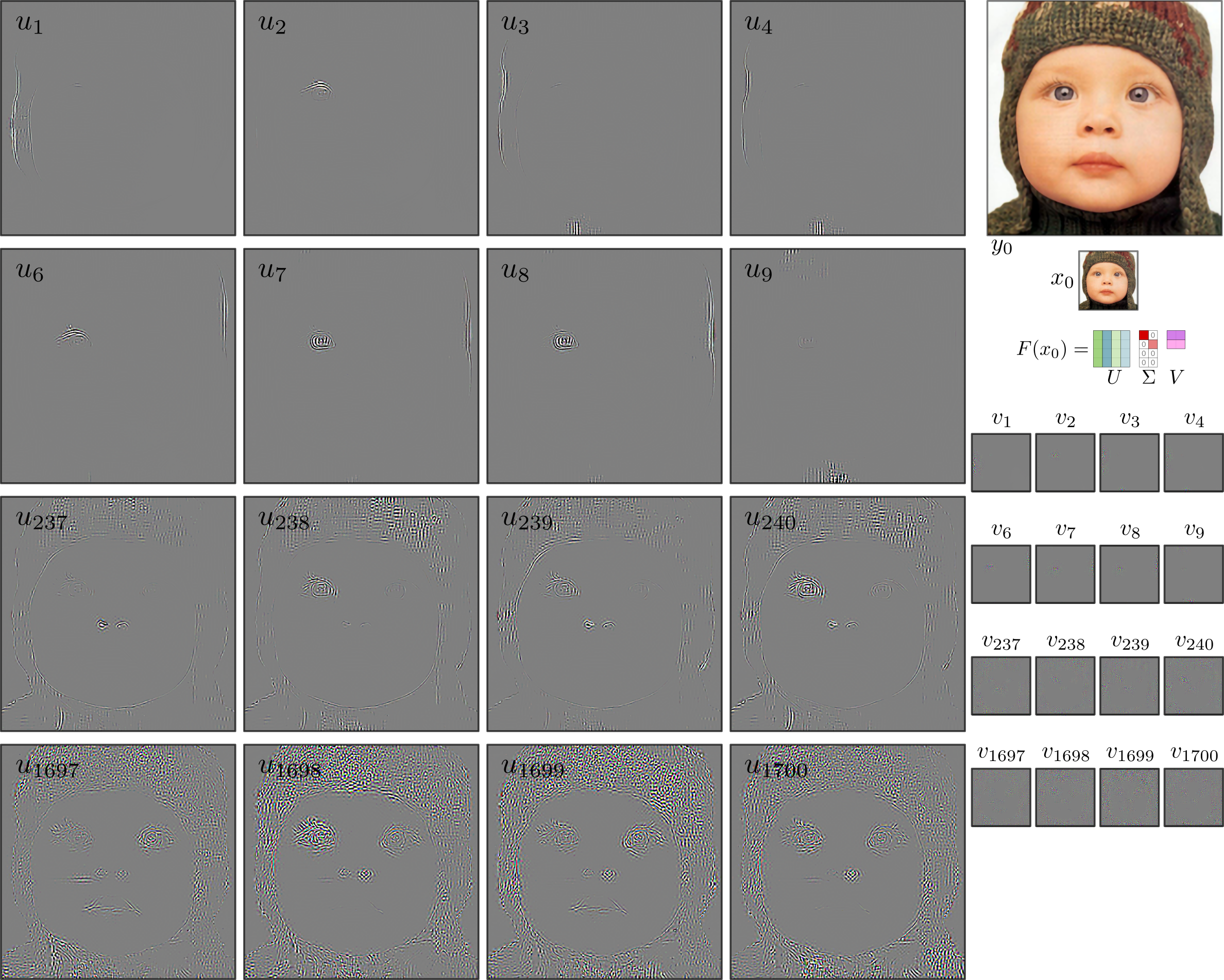}
    \caption{Results of the Singular Value decomposition of a linear interpreter applied on EDSR\cite{Lim_2017_CVPR_Workshops} $4\times$ super--resolution method. The basis used by EDSR is spatially localized, oriented and bandpass, comparable to wavelet basis\cite{olshausen1996emergence}, and similar to the receptive fields of simple cells in mammalian primary visual cortex. The Eigen--inputs/outputs for largest singular values capture the objects with largest receptive fields, indicating strong knowledge of the geometry of the image.}
\label{fig:SR_eigenvectors}
\end{figure*}
\textbf{Projective/receptive filters}:

Supplementary material in section \ref{app:links} includes a live demonstration, showing rows and columns of the linear interpreter for the upscaling methods: Bicubic, $4$--layers PixelShuffle\footnote{Model obtained by running a PyTorch tutorial from \url{https://github.com/pytorch/examples/tree/master/super_resolution}}, and EDSR\cite{Lim_2017_CVPR_Workshops}. In Figure \ref{fig:FilterMatrix_SR} we show snapshots of the demonstration. The filter matrix $F(x_0)$ for SR methods is not square and has a vertical shape. For every pixel in the input domain (small resolution) there is a column, representing the projective filter. We implement and analyze a Bicubic upscaler for two reasons: first, it helps to verify the implementation of our analysis; and second, to take it as a reference for interpretation. The demonstration let users move around an image and inspect all the filter matrix's rows and columns. It is the equivalent to materialize the matrix $F(x_0)$, except that we do not keep the matrix in memory. For these examples we precomputed all rows and columns and save them as image files. For the largest and slowest model, EDSR, we can compute more than $2$ rows and column images per second on a Titan X GPU (12GB). Then, we use a modern browser that displays the diagram and loads the row/column images corresponding to the location in the image.

For SR methods we observe that \textbf{filter coefficients are sparse} since the image outside the zooming window is mostly full of zeros. The \textbf{coefficients are concentrated around the pixel location}, as expected, since interpolation must give preference to the current pixel location and use its neighbors to improve it.

The demonstration shows that for good models, like EDSR, a user can guess the location in the image just by looking at the projective filters (columns). In layman's terms:
\begin{quote}
    \textbf{\emph{Inspecting SR projective filter coefficients feels like walking through the image with a flashlight.}}
\end{quote}
This observation offers a simple check to verify that the model has learned the geometry of images. On the other end, bicubic would make a user feel blind since it is completely space invariant; and the $4$--layers PixelShuffle model would make users feel confused on the location because the geometry is not clearly revealed in the projective filters.

\bigskip
\textbf{More eigen--inputs/outputs}:

The images of eigen--input/outputs in the main text contain zooms that cover certain areas of the image, often full of zeros. In Figure \ref{fig:SR_eigenvectors} we show more eigen--input/outputs without zooming. Here, we observe that eigen--input/outputs are sparse and capture few or single features of the image (e.g. left eye, right eye, etc.) for the largest singular values. The images of eigen--input/outputs for larger singular values contain higher--frequencies and cover larger areas.

\subsection{Image--to--Image Translation (I2I)}
\label{app:I2I}
\begin{figure*}[h!]
    \centering
    \framebox{\includegraphics[width=.3\linewidth]{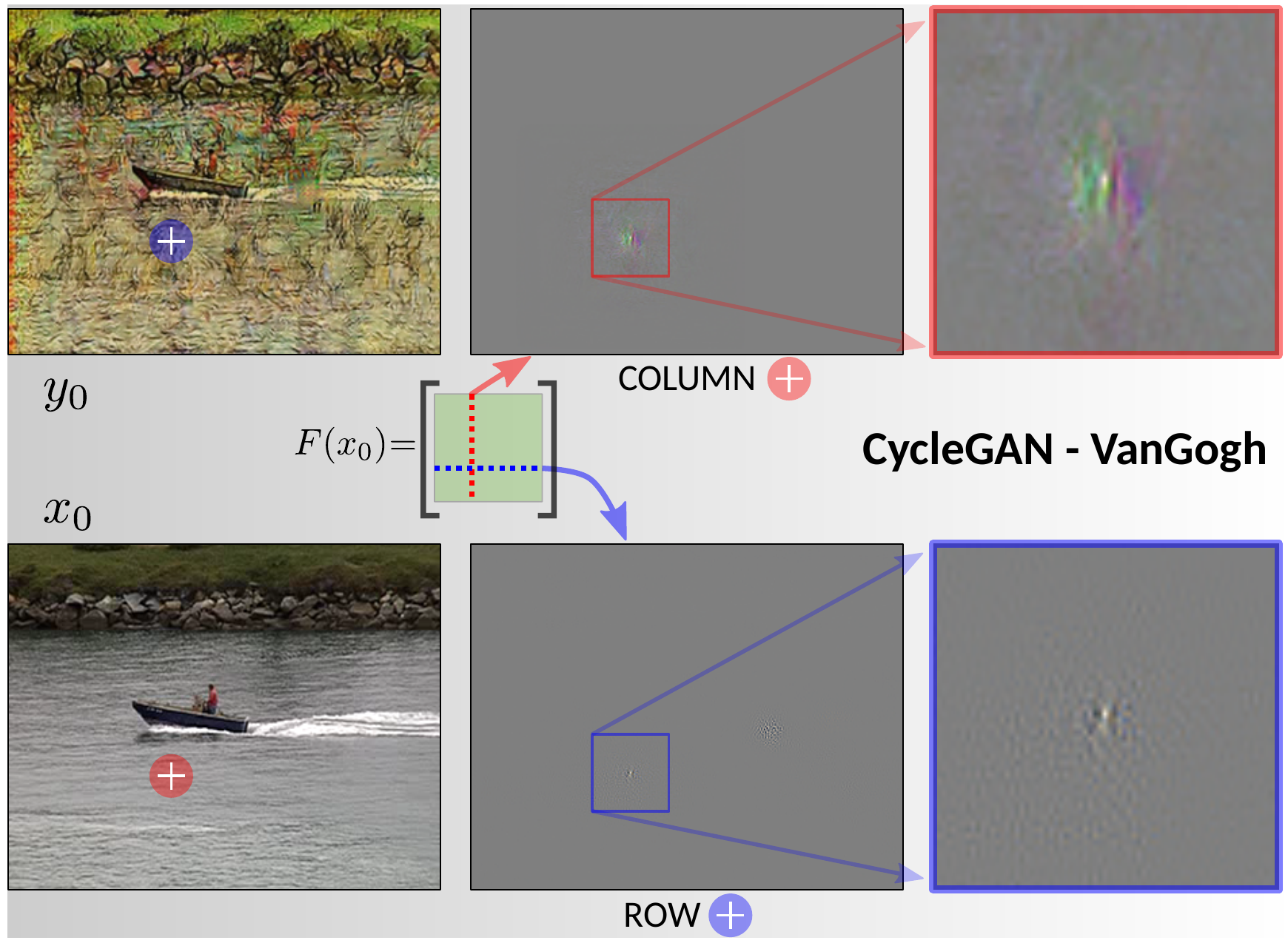}}
    \framebox{\includegraphics[width=.3\linewidth]{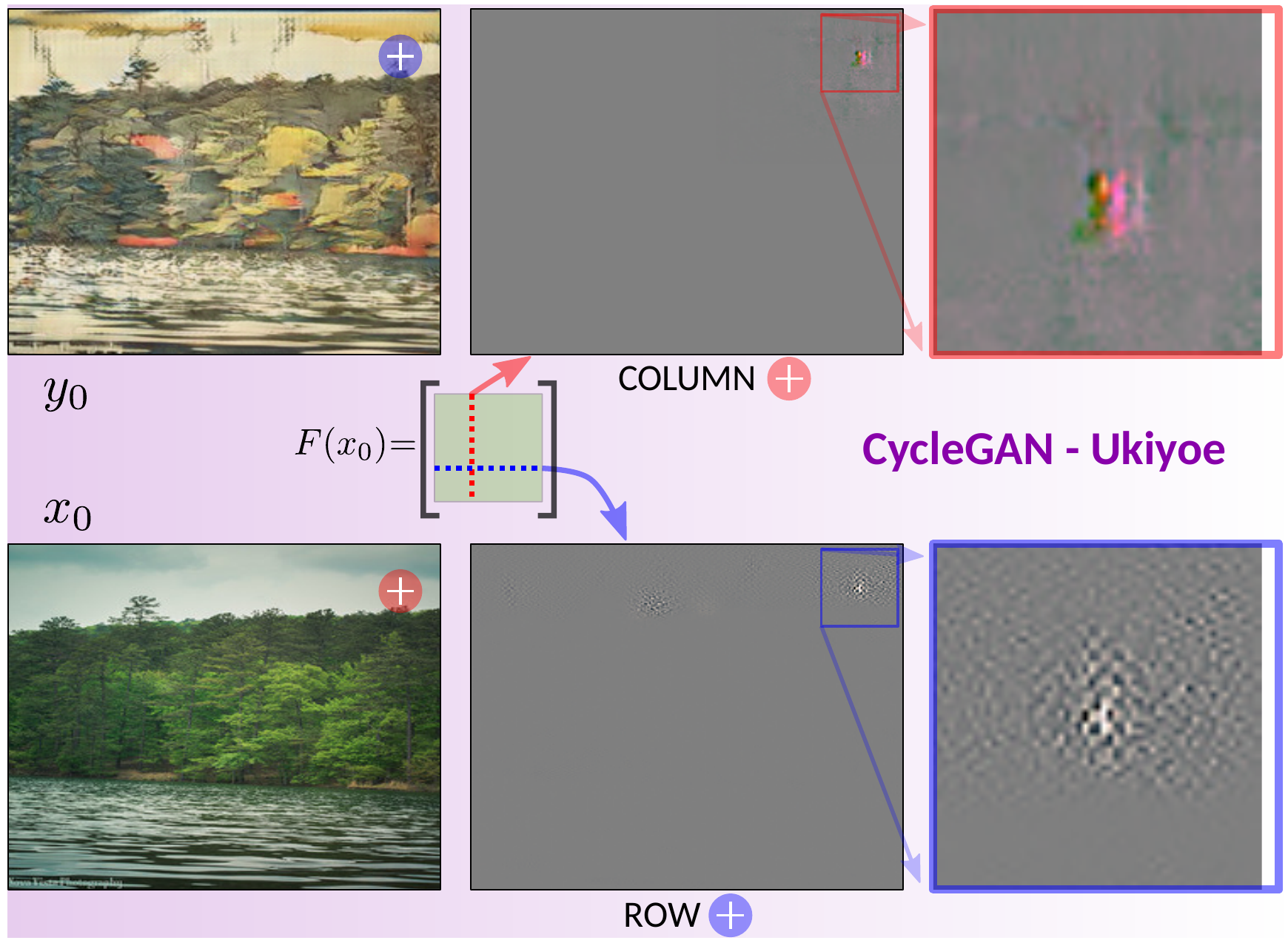}}
    \framebox{\includegraphics[width=.3\linewidth]{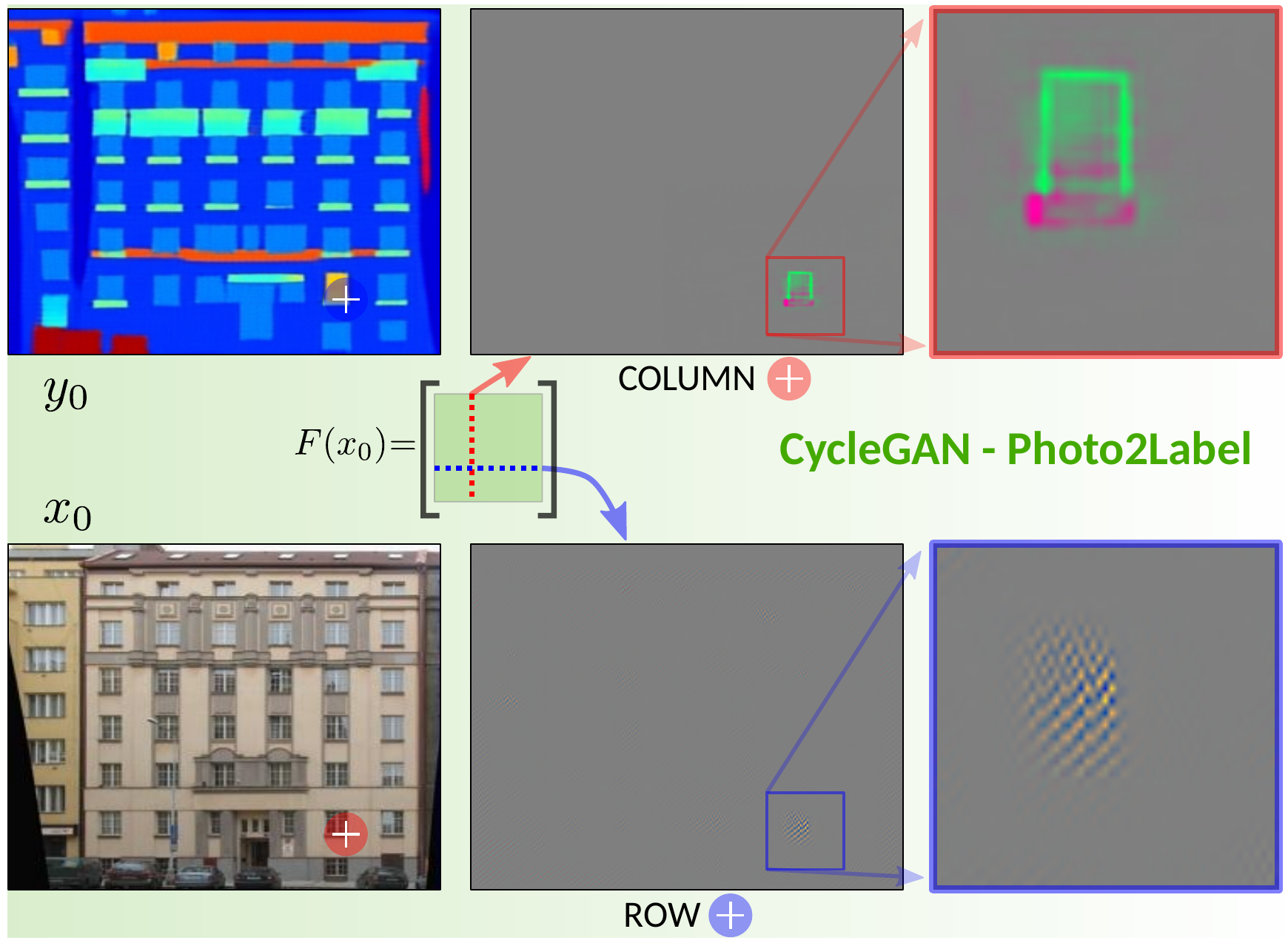}}
    \caption{Screenshots of live demonstration showing columns and rows for the linear interpreter of image--to--image translation systems. The demonstration reveals strong presence of off-diagonals in the filter matrix. This means that CycleGAN chooses certain areas in a given image to copy, move and generate textures.}
\label{fig:FilterMatrix_I2I}
\end{figure*}
\begin{figure*}[h!]
    \centering
    \includegraphics[width=.87\linewidth]{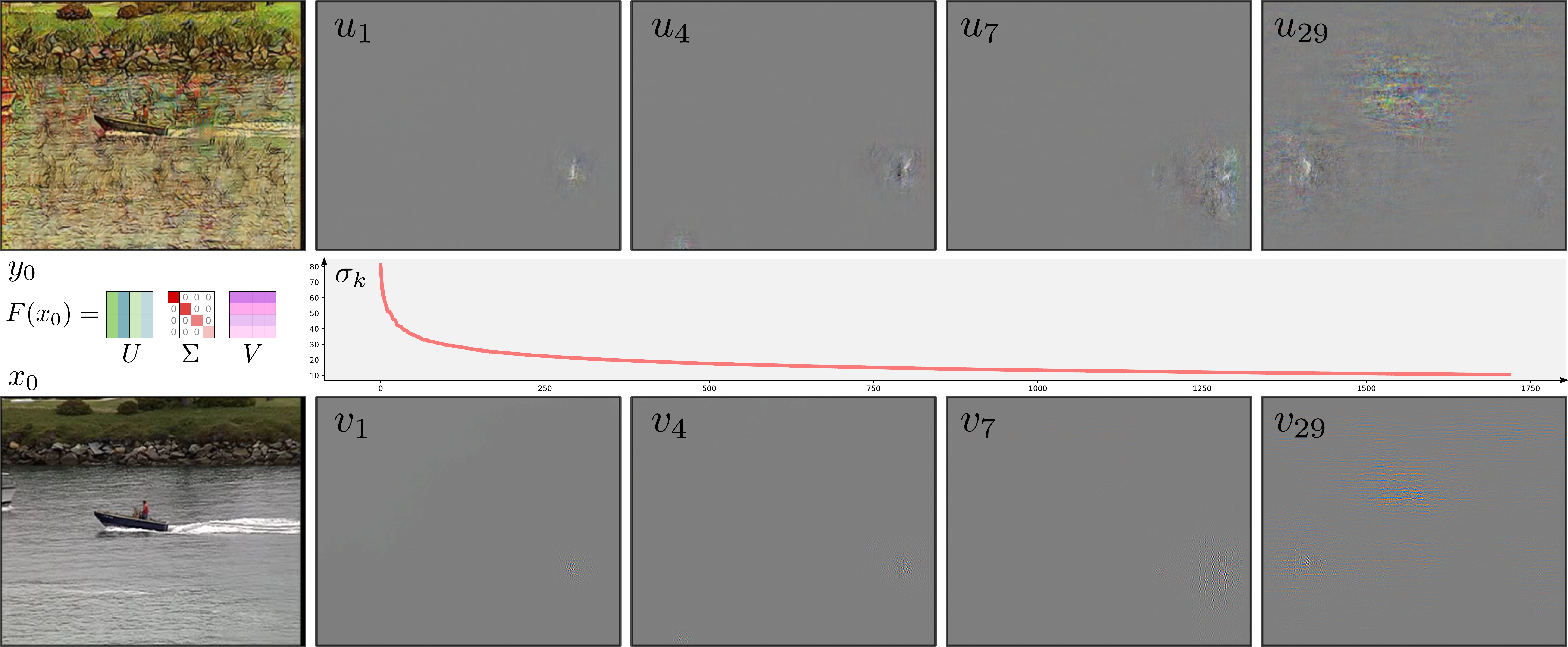}
    \caption{Results of the Singular Value decomposition of a linear interpreter applied on CycleGAN\cite{CycleGAN2017}--VanGogh I2I network model. Eigen--outputs for large singular values reveal the areas with largest contributions. Compared to SR eigen--decompositions, the basis is also spatially localized, oriented and bandpass, comparable to wavelet basis\cite{olshausen1996emergence,SMallat_1998a}. But we observe that I2I eigen--decomposition is much less sparse than in SR, indicating a more global strategy to solve the problem.}
\label{fig:I2I_svd}
\end{figure*}
\begin{figure*}[h!]
    \centering
    \includegraphics[width=.76\linewidth]{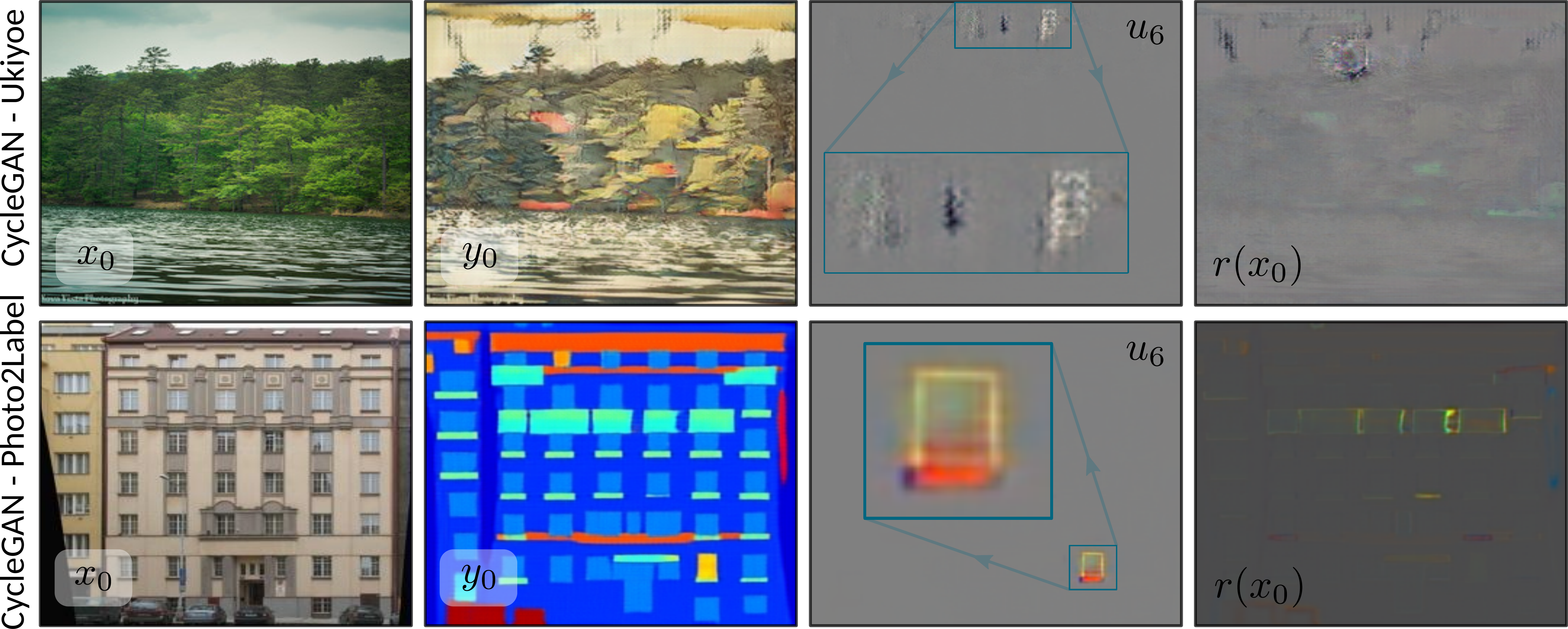}
    \caption{The SVD of I2I models shows how the network focuses on particular styles. Residuals in I2I contribute more than in SR problems. Eigen--outputs for large singular values help to identify the areas with largest contributions. In Ukiyoe style, the eigen--output $u_6$ shows an area originally empty in the input, where a new texture has been created. In Photo--to--Label, the eigen--output $u_6$ shows the creation of a template window segment.}
\label{fig:I2I_compare}
\end{figure*}
\textbf{Projective/receptive filters}:

Supplementary material in section \ref{app:links} includes a live demonstration, showing rows and columns of the linear interpreter for CycleGAN\cite{CycleGAN2017} architecture and different pre--trained models\footnote{CycleGAN models downloaded from \url{http://efrosgans.eecs.berkeley.edu/cyclegan/pretrained_models}}. In Figure \ref{fig:FilterMatrix_I2I} we show some snapshots of the live demonstration.

The filter matrix $F(x_0)$ for I2I methods is square. Here, we observe that \textbf{filters coefficients in I2I are less sparse than those in SR methods}. In areas where the content does not show strong changes we observe delta-type filter coefficients centered in the diagonal. In areas where the content is converted to a cartoon--style flat color (e.g. blue background in Photo--to--Label) the input is largely ignored or spread around a large area. In other areas the coefficients are strong around the diagonal but often include strong off--diagonal components (see Figure \ref{fig:FilterMatrix_I2I}). We observe strong off--diagonal components in the columns, suggesting that the network is using both pixel values in current location, as well as values from other regions of the input image, in order to obtain the output. We also see strong off--diagonal components in the rows, suggesting that the network is using the results of the current location somewhere else in the image. The CycleGAN\cite{CycleGAN2017} architecture can achieve this easily by using instance--normalization layers that make use of global features (image mean and variance).

As opposed to good SR methods, when using painting styles (VanGogh and Ukiyoe) the projective filters (columns) do not follow the geometry of the image and do not easily reveal the location in the image. In the case of the Photo--to--Label model we can guess the content and location because we see windows with strong neon--style colors. \textbf{The filter coefficients in painting styles seem to focus more on textures and color}.

In the case of Photo--to--Label style, we do not observe a peak in the diagonal elements (the location of the current pixel) as seen before in SR methods and painting styles when content is preserved. Instead, we see the shape of windows turning on and off. We believe that this is caused by the nature of the problem, that is basically trying to perform segmentation. \textbf{The Photo--to--Label filter matrix works like a detection system that creates template boxes in the output}. The background blue color indicates that a segment has not been detected. The on--and--off effect suggests that a new segment has been found (e.g. an eave, a window, a door, etc). The receptive filter (rows) resembles a Gabor--like template matching filter. The projective filters (columns) show how single pixels are assigned to a whole segment in the output. For this problem, templates are simple and the network is able to create them. This is much simpler than creating templates for image classes in ImageNet where we did not observe the network following the same strategy.

\bigskip
\textbf{Eigen--inputs/outputs}:

In Figure \ref{fig:I2I_svd} we observe how CycleGAN's eigen--decompositions show some similar patterns compared to SR models. Namely, eigen--inputs/outputs are localized for large singular values and cover larger areas for smaller singular values. Also, eigen--inputs contain high frequency stimulus that are translated into colorful textures (VanGogh and Ukiyoe styles) or template boxes (Photo--to--Label style) in their correspondent eigen--outputs.

Other patterns are clearly different. Namely, the first eigen--inputs/outputs cover larger areas than SR models, and they focus more on color, capturing some of the VanGogh style used in Figure \ref{fig:I2I_svd}. The content in eigen--outputs capture more textures, compared to SR models that focus more on curves and edges.

In Figure \ref{fig:I2I_compare} we observe that residuals in CycleGAN models are larger than residuals observed in SR. The eigen--outputs of different styles show a clear focus of the network in generating the colors and objects of the target style. We can conclude that an SVD analysis helps to interpret a network by showing how they focus on their tasks. This is, geometric shapes for SR and texture/color styles for I2I.

\subsection{Demonstrations}
\label{app:links}

\textbf{Video material}: The following videos are included as supplementary material:
\begin{itemize}
    \item \href{https://www.dropbox.com/s/s7223ml92jvlz7y/FilterMatrix_SR.mp4}{FilterMatrix\_SR.mp4}
    \item \href{https://www.dropbox.com/s/63jcvj17kxp61cj/FilterMatrix_I2I.mp4}{FilterMatrix\_I2I.mp4}
\end{itemize}
Both videos include an English subtitle track embedded in the MP4 containers. We hope that these comments can help viewers to better understand the results of the analysis. The subtitle's font and size are controlled by the video player and can sometimes obstruct information in the video frames. Please feel free to enable/disable the subtitle track to better appreciate the results of the demonstration.

\textbf{Interactive material}: The interactive demonstrations can be downloaded from:
\begin{itemize}
    \item \href{https://www.dropbox.com/s/rn086hyh13pvuvt/Bicubic4x.zip}{Bicubic4x.zip (17 MB)}
    \item \href{https://www.dropbox.com/s/zum5hmtl9smo908/4L-PixelShuffle.zip}{4L-PixelShuffle.zip (17 MB)}
    \item \href{https://www.dropbox.com/s/3q6g3hnzmbleuh0/EDSR4x.zip}{EDSR4x.zip (953 MB)}
    \item \href{https://www.dropbox.com/s/7gbrb2vcukx5i18/CycleGAN-VanGogh.zip}{CycleGAN-VanGogh (7.5 GB)}
    \item \href{https://www.dropbox.com/s/zhpi9hxvao2xbyg/CycleGAN-Ukiyoe.zip}{CycleGAN-Ukiyoe (4.4 GB)}
    \item \href{https://www.dropbox.com/s/547y44ao5hccjbk/CycleGAN-Photo2Label.zip}{CycleGAN-Photo2Label (4.0 GB)}
\end{itemize}
Please note that large file sizes (mostly I2I) are due to the fact that we recorded all rows and columns using lossless compression to avoid misinterpretations.

{\small
\bibliographystyle{ieee_fullname}
\bibliography{bibliography}
}

\end{document}